%% file: main.tex
\documentclass[10pt,twocolumn,letterpaper]{article}
\usepackage{cvpr}              % To produce the CAMERA-READY version
\input{preamble}

\definecolor{cvprblue}{rgb}{0.21,0.49,0.74}
\usepackage[pagebackref,breaklinks,colorlinks,allcolors=cvprblue]{hyperref}
\usepackage{placeins} % debug

\def\paperID{XXXX} 
\def\confName{CVPR}
\def\confYear{2026}

\title{%
  V$^{2}$-SAM: Marrying SAM2 with Multi-Prompt Experts for \\ Cross-View Object Correspondence
}

\author{Jiancheng Pan$^{1,2,*}$  \quad Runze Wang$^{3,}$\thanks{These authors have equal contributions.} \quad Tianwen Qian$^4$ \quad Mohammad Mahdi$^1$  \quad Yanwei Fu$^3$ \\ Xiangyang Xue$^3$ \quad Xiaomeng Huang$^2$  \quad Luc Van Gool$^1$ \quad Danda Pani Paudel$^1$ \quad Yuqian Fu$^1$\thanks{Corresponding author.} \\
$^1$INSAIT, Sofia University ``St. Kliment Ohridski”, $^2$Tsinghua University\\$^3$Fudan University, $^4$East China Normal University\\
{\tt\small jiancheng.pan.plus@gmail.com}, 
{\tt\small wangrz24@m.fudan.edu.cn}, 
{\tt\small yuqian.fu.ai@gmail.com}}

\usepackage{amssymb}

\usepackage[table]{xcolor}  % 放在导言区
\definecolor{vpblue}{HTML}{3D82C7}     % VP (Dense)
\definecolor{sparseorange}{HTML}{E66B2D}% Sparse
\definecolor{fusiongreen}{HTML}{4CB944}% Fusion


\begin{document}
\maketitle

\input{sec/0_abstract}
\input{sec/1_intro}
\input{sec/2_related_works}

\input{sec/3_methods}

\input{sec/4_results}

\input{sec/5_conclusions}
% \input{sec/6_acknowledgements}

% \FloatBarrier % debug
% \clearpage
% \clearpage
{
    \small
    \bibliographystyle{ieeenat_fullname}
    \bibliography{main}
}

\input{sec/7_suppl}


\end{document}

%% file: preamble.tex
\newcommand{\wrz}[1]{\begin{color}{black}#1\end{color}}

\usepackage[accsupp]{axessibility}
\usepackage{amssymb}
\usepackage{pifont}
\usepackage{multirow}
\usepackage{commath}
\usepackage{makecell}
\usepackage{pifont}
\usepackage[table]{xcolor}
\usepackage{appendix}
\usepackage{xr-hyper}

\makeatletter
\newcommand*{\addFileDependency}[1]{
  \typeout{(#1)}
  \@addtofilelist{#1}
  \IfFileExists{#1}{}{\typeout{No file #1.}}
}
\newcommand*{\newbibstartnumber}[1]{%
  \apptocmd{\thebibliography}{%
    \global\c@NAT@ctr #1\relax
    \addtocounter{NAT@ctr}{-1}%
  }{}{}%
}
\makeatother

%% file: sec/0_abstract.tex
\begin{abstract}

Cross-view object correspondence, exemplified by the representative task of ego–exo object correspondence, aims to establish consistent associations of the same object across different viewpoints (e.g., ego-centric and exo-centric). This task poses significant challenges due to drastic viewpoint and appearance variations, making existing segmentation models, such as SAM2, non-trivial to apply directly. To address this, we present V$^{2}$-SAM, a unified cross-view object correspondence framework that adapts SAM2 from single-view segmentation to cross-view correspondence through two complementary prompt generators. Specifically, the Cross-View Anchor Prompt Generator (V$^{2}$-Anchor), built upon DINOv3 features, establishes geometry-aware correspondences and, for the first time, unlocks coordinate-based prompting for SAM2 in cross-view scenarios, while the Cross-View Visual Prompt Generator (V$^{2}$-Visual) enhances appearance-guided cues via a novel visual prompt matcher that aligns ego–exo representations from both feature and structural perspectives. To effectively exploit the strengths of both prompts, we further adopt a multi-expert design and introduce a Post-hoc Cyclic Consistency Selector (PCCS) that adaptively selects the most reliable expert based on cyclic consistency. Extensive experiments validate the effectiveness of V$^{2}$-SAM, achieving new state-of-the-art performance on Ego-Exo4D (Ego–Exo object correspondence), DAVIS-2017 (video object tracking), and HANDAL-X (robotic-ready cross-view correspondence). Codes and Models will be released at \href{https://jianchengpan.space/projects/V2-SAM/}{\textcolor{pink}{V$^{2}$-SAM}}.

\end{abstract}

%% file: sec/1_intro.tex
    \addtocontents{toc}{\protect\setcounter{tocdepth}{-1}}

\section{Introduction}
\label{sec:intro}

%p1: task of cross-view correspondence --> ego-exo 
Cross-view object correspondence~\cite{fu2024objectrelator} aims to associate object identities across viewpoints by segmenting the same object(s) in a target view, given their mask annotation in a query view. This ability is fundamental for multi-view scene understanding~\cite{jaritz2019multi,yu2025inst3d,miao2024scenegraphloccrossmodalcoarsevisual,Liu_Sun_Xie_Li_Li_Zhang_2025}, video perception~\cite{zhang2025vividface, fu2020depth, miao2024volumetricsemanticallyconsistent3d, feng2025pmq, peng2024towards, peng2024referring, zhang2025egonight}, and embodied AI~\cite{9687596,Savva_2019_ICCV, li2025segment, li2025clivis}. Among various cross-view scenarios, ego–exo object correspondence~\cite{grauman2024ego,mur2025mama,liao2025domr,hu2025robust} stands out as a particularly representative and challenging case. In this setting, the same object is simultaneously observed from a moving first-person (ego-centric) camera and a static third-person (exo-centric) camera, leading to drastic viewpoint and appearance discrepancies, complex background clutter, dynamic motions, and occlusions.

\begin{figure}[t]
    \centering
    \includegraphics[width=0.98\linewidth]{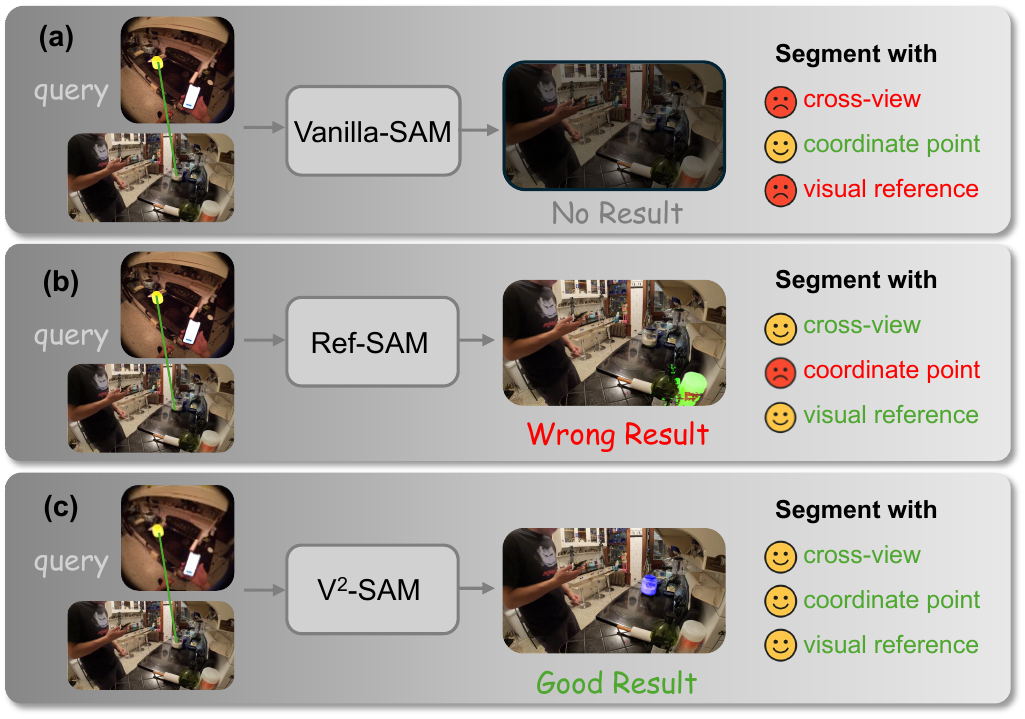}
    \caption{Comparison of SAM variants in segmentation capability. Our proposed V²-SAM supports coordinate-point and visual-reference prompts for cross-view segmentation.}
    \label{fig:teaser}
\end{figure}

%P2. From single-view segmentation to cross-view segmentation: the gap and limitations
Recent advances in single-view segmentation have been driven by powerful foundation models such as SAM2~\cite{ravi2024sam2segmentimages} (as Fig.~\ref{fig:teaser}(a)), which exhibit remarkable mask generation capability given well-localized prompts. However, extending such models to the cross-view setting is highly non-trivial. SAM2 depends on spatially grounded prompts (e.g., mask coordinates or bounding boxes) to condition its decoder, but in cross-view scenarios, the target object’s position can vary drastically across views, making direct position-based prompting infeasible. While referring-based extensions~\cite{ravi2024sam2segmentimages,bi2025virefsam} (as Fig.~\ref{fig:teaser}(b) Ref-SAM) replace explicit positional prompts with visually guided prompt generators, they still face two major limitations: 
1) they often fail under severe appearance changes across views; and
2) removing spatial prompts undermines SAM2’s inherent strength in localization.
The above-mentioned challenges motivate us to explore two central questions: \ding{172} Can we unlock SAM2’s spatial prompting capability in cross-view scenarios? \ding{173} If so, can spatial and visual prompts complement each other to further enhance cross-view segmentation performance?

%p4: how to tackle the 1) challenge ; and 2) challenge
To address \ding{172}, we propose a novel \textit{Cross-View Anchor Prompt Generator (V$^{2}$-Anchor)} that restores SAM2’s ability to localize objects across views. Specifically, we leverage DINOv3’s geometry-aware feature space to identify corresponding object regions between ego- and exo-centric images. This is achieved through cross-view feature matching followed by point stratification to establish reliable correspondences. To suppress noisy correspondences and avoid confusion with nearby distractors in cluttered scenes, outliers are removed and a centroid point is selected from the remaining matches, yielding a high-confidence coordinate in the target view. In this way, we effectively \textit{unlock coordinate-based prompting} for SAM2 in cross-view segmentation, particularly under the ego–exo setting. As for \ding{173}, inspired by prior visual-referring segmentation models~\cite{Jing_2021_CVPR,sun2024vrp,bi2025virefsam}, we develop the \textit{Cross-View Visual Prompt Generator (V$^{2}$-Visual)}, which introduces a novel \textit{Visual Prompt Matcher} to align cross-view object representations from both feature and structural perspectives. Empirically, we observe that V$^{2}$-Anchor excels at determining \textit{where it is}, while V$^{2}$-Visual is more effective in identifying \textit{what it looks like}. Motivated by their complementary strengths, we design a multi-prompt expert framework that exploits both prompt types. Inspired by the Mixture-of-Experts ~\cite{he2021fastmoefastmixtureofexperttraining} (MoE) paradigm~\cite{6215056,NEURIPS2022_2f00ecd7}, we define each expert as a combination of a prompt generator and a SAM2-based mask decoder, and train three experts: one specialized in spatial prompts (Anchor Expert), one in visual prompts (Visual Expert), and one in fused spatial–visual prompts (Fusion Expert). Finally, we introduce a \textit{Post-hoc Cyclic Consistency Selector (PCCS)} that adaptively selects the optimal expert for each object based on cross-view cyclic consistency of the predicted masks.

%p5, formally, and exps
Formally, by integrating our V$^{2}$-Anchor, V$^{2}$-Visual, multi-expert training, and the PCCS into the vanilla SAM2, we construct \textbf{V$^{2}$-SAM} (as Fig.~\ref{fig:teaser}(c)), a unified cross-view object segmentation framework that effectively mitigates the inherent challenges introduced by drastic viewpoint and appearance variations discussed earlier. We conduct extensive experiments across a range of cross-view correspondence benchmarks, including Ego-Exo4D (ego–exo object correspondence), DAVIS-2017 (video object tracking), and HANDAL-X (robotic-ready cross-view correspondence). Results show that V$^{2}$-SAM achieves state-of-the-art (SOTA) performance on all three benchmarks, consistently outperforming strong baselines and existing competitors.
%p6, main contri
Our main contributions are summarized as follows:
\begin{itemize}
\item \textbf{Unified framework:} We propose V$^{2}$-SAM, \wrz{the first unified framework to architecturally adapt SAM2 to cross-view object correspondence, especially for the challenging ego–exo task.} 
\item \textbf{Cross-view prompt generators:} V$^{2}$-Anchor, for the first time, enables coordinate-based prompting for SAM2 in cross-view scenarios; V$^{2}$-Visual enhances appearance-guided cues via a novel visual prompt matcher.
\item \textbf{Expert integration and selection:} A multi-prompt expert framework with the novel PCCS module adaptively selects the most reliable expert for each instance.
\item \textbf{Extensive validation:} The effectiveness of V$^{2}$-SAM is validated on Ego-Exo4D, DAVIS-2017, and HANDAL-X, achieving new SOTA performance.
\end{itemize}

% We propose V$^{2}$-SAM, the first framework that adapts SAM2 to cross-view object correspondence, especially for the challenging ego–exo task. 

%% file: sec/2_related_works.tex
\section{Related Works}
\label{sec:related}

\begin{figure*}[t]
    \centering
    \includegraphics[width=0.9\linewidth]{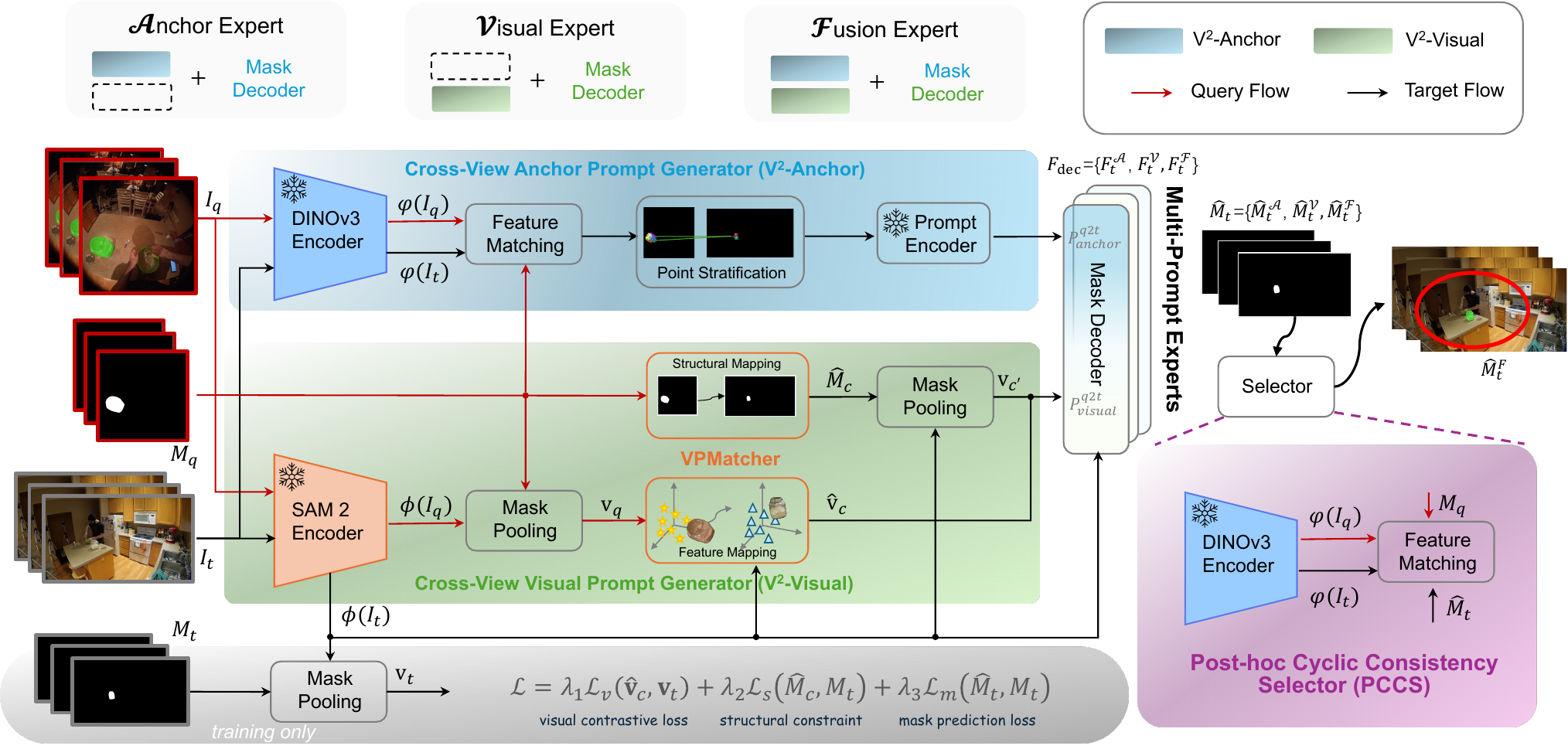}
    % \caption{V$^{2}$-SAM framework. It introduces V$^{2}$-Anchor for coordinate-based cross-view prompting, V$^{2}$-Visual for enhanced appearance-guided visual matching, and a multi-prompt expert framework equipped with the PCCS module for adaptive expert selection.} 
    \caption{\wrz{Overview of V$^{2}$-SAM. Given a query–target image pair $(I_q, I_t)$ and the query object mask $M_q$, we generate two cross-view prompts: a geometry-guided cue $P^{q2t}_{anchor}$ via V$^{2}$-Anchor (blue) and an appearance-guided cue $P^{q2t}_{visual}$ via V$^{2}$-Visual (green). Each prompt is fed into its corresponding mask decoder to produce candidate target masks, and the Post-hoc Cyclic Consistency Selector (PCCS) adaptively selects the most reliable prediction as the final output. }} 
    \label{fig:architecture}
\end{figure*}

% \noindent \textbf{Cross-View Object Correspondence.}
% Cross-view object correspondence, a key problem in ego–exo perception~\cite{thatipelli2025egocentric,mahdi2025exo2egosyn}, involves taking object queries from one view (e.g., ego-centric) and predicting the corresponding object masks in another (e.g., exo-centric). Baade et al.~\cite{baade2025self} first introduced Predictive Cycle Consistency for learning object correspondence between extremely disjoint views of a scene without the need for paired segmentation data. O-MaMa~\cite{mur2025mama} and DOMR~\cite{liao2025domr} achieved cross-view segmentation by treating it as a mask matching task. However, these methods rely heavily on the availability of a robust segmentation model and often lack the capacity for learning and generalization. To bridge the domain gap between egocentric and exocentric perspectives, Fu et al.~\cite{fu2024objectrelator, fu2025cross} propose ObjectRelator to achieve end-to-end feature-guided ego-exo cross-view object segmentation. And Hu et al.~\cite{hu2025robust} propose a robust ego–exo association framework that incorporates long-term memory to achieve reliable video-level matching across extended sequences and significant viewpoint variations. Nonetheless, this approach still provides limited memory optimization across frames, making fine-grained ego–exo frame-level alignment a persistent challenge.

\noindent \textbf{Cross-View Object Correspondence.}
\wrz{Cross-view object correspondence, a key problem in ego–exo perception~\cite{thatipelli2025egocentric,mahdi2025exo2egosyn}, involves taking object queries from one view and predicting the corresponding object masks in another. Existing methods can be broadly grouped into two paradigms.
\textit{Matching-based methods} formulate the task as post-hoc matching. Baade et al.~\cite{baade2025self} first introduced Predictive Cycle Consistency to construct matched object pairs across views for training. More recently, O-MaMa~\cite{mur2025mama} and DOMR~\cite{liao2025domr} achieve cross-view segmentation by matching mask proposals from an external segmentation model, thereby limiting their learning and generalization capacity.
\textit{Learnable model-based methods}, in contrast, directly learn cross-view object segmentation. Fu et al.~\cite{fu2024objectrelator, fu2025cross} proposed ObjectRelator, an end-to-end framework for cross-view segmentation using visual and textual condition. 
To improve robustness and generalization, our method follows the learnable model-based paradigm. Unlike ObjectRelator, we introduce coordinate-based cues to unlock SAM2 for cross-view correspondence, a largely unexplored direction. Moreover, cross-view correspondence varies significantly across scene conditions. We further develop a multi-expert framework to effectively exploit the complementary strengths of both prompts.}

\noindent \textbf{Visual Reference-Guided Segmentation.} 
Recent studies have extended the prompt-driven paradigm of the Segment Anything Model (SAM)~\cite{kirillov2023segment} toward visual reference-guided segmentatio~\cite{miao2025langhopslanguagegroundedhierarchical}, where visual-reference images serve as prompts for segmentation. VRP-SAM~\cite{sun2024vrp} and ViRefSAM~\cite{bi2025virefsam} utilize reference-image encoders to guide SAM for class-specific or domain-adaptive segmentation. Meanwhile, some studies have attempted to combine visual and textual cues for joint prompt segmentation, e.g., VLP-SAM~\cite{sakurai2025vision}, which combines visual and textual cues to improve few-shot generalization. Unlike these methods, which mainly focus on intra-view visual prompting, our work explores cross-view correspondence, emphasizing the integration of spatial and appearance cues to enable consistent segmentation. This perspective extends visual reference-guided segmentation toward more general and embodied perception scenarios.

%% file: sec/3_methods.tex
\section{Methodology}\label{sec:methods}
\noindent\textbf{Task Formulation.} Formally, given two temporally aligned image/video from distinct views, a query view $I_{q}$/$V_{q}$ and a target view $I_{t}$/$V_{t}$, and a query object mask(s) $M_{q}$ in the query view, the objective is to predict the corresponding object mask $\hat{M}_{t}$ in the target view. Ego–Exo object correspondence serves as a representative case of this formulation, in which the two views come from egocentric and exocentric cameras, respectively. Depending on the transfer direction, the task can be defined as an Ego2Exo or an Exo2Ego subtask. In all cases, the model relies solely on visual information without access to camera poses, semantic labels, or explicit 3D geometry.

% \noindent\textbf{SAM2 Baseline.} 
% SAM2~\cite{ravi2024sam2segmentimages} is one of the most powerful segmentation models to date, demonstrating outstanding generalization ability through visual prompting. It consists of a \textit{Image Encoder} for visual feature extraction, a \textit{Prompt Encoder} for embedding points, boxes, or masks, and a \textit{Mask Decoder} that predicts object masks using features from both encoders. A memory mechanism further enables temporal mask propagation across video frames. However, since all prompts are defined within the coordinate system of the target image, SAM2 is inherently non-trivial to adapt to cross-view scenarios, where object position, scale, shape, and appearance often vary drastically across views.

\subsection{Overview of Proposed V$^2$-SAM}\label{sec:overview}

% Our framework is built upon SAM2, retaining its core components. Specifically, we keep the \textit{SAM2 Encoder} $\phi(\cdot)$, \textit{Prompt Encoder}, and \textit{Mask Decoder}, while discarding memory-related modules to focus on frame-level correspondence. This design allows the framework to generalize seamlessly across both cross-view image and video tasks.  
% To unlock SAM2’s potential for cross-view object correspondence, we introduce four novel modules:  
% 1) a \textit{Cross-View Anchor Prompt Generator (V$^{2}$-Anchor)} that transfers the query mask’s spatial information to the target view using DINOv3 $\varphi(\cdot)$’s geometry-aware feature space, for the first time enabling coordinate-based prompting across views;  
% 2) a \textit{Cross-View Visual Prompt Generator (V$^{2}$-Visual)} that leverages object appearance cues and refines them through a learnable mapping between views;  
% 3) a \textit{Multi-Expert Training} mechanism that jointly learns spatial, visual, and fused experts for complementary reasoning; and  
% 4) a \textit{Post-hoc Cyclic Consistency Selector (PCCS)} that adaptively selects the most reliable expert at inference based on cross-view mask consistency. Together, these components form our V$^2$-SAM, a unified segmentation framework that bridges spatial alignment and semantic association across drastically different viewpoints.

SAM2~\cite{ravi2024sam2segmentimages} is one of the most powerful segmentation models to date, demonstrating outstanding generalization ability through visual prompting. Our framework is built upon SAM2, retaining its core components. Specifically, we keep the \textit{SAM2 Encoder} $\phi(\cdot)$, \textit{Prompt Encoder}, and \textit{Mask Decoder}, while discarding memory-related modules to focus on frame-level correspondence. This design allows the framework to generalize seamlessly across both cross-view image and video tasks.

The overall workflow of V$^{2}$-SAM is illustrated in Fig.~\ref{fig:architecture}. Given a query–target image pair $(I_q, I_t)$ and the query object mask $M_q$, the model first employs our two prompt generators, i.e., \textit{Cross-View Anchor Prompt Generator} (V$^{2}$-Anchor) and \textit{Cross-View Visual Prompt Generator} (V$^{2}$-Visual), to transform the object information from the query view to the target view. The resulting cross-view prompts, denoted as $P^{q2t}_{anchor}$ and $P^{q2t}_{visual}$, encode geometry- and appearance-guided cues, respectively.  
Breiefly, the V$^{2}$-Anchor branch operates on DINOv3 features to establish geometric correspondences between $(I_q, I_t)$. It identifies reliable point-level matches, refines them via stratified filtering, and converts them into coordinate-based prompts $P^{q2t}_{anchor}$ that localize the object in the target view. In parallel, the \textit{V$^{2}$-Visual} branch extracts region-level representations by applying mask pooling on SAM2 features within the query mask $M_q$. These representations are then transferred to the target view through a learnable Visual Prompt Matcher (VPMatcher), producing the appearance-guided prompt $P^{q2t}_{visual}$.  
With these two complementary prompts, three experts are constructed and trained: the \textit{Anchor Expert}, the \textit{Visual Expert}, and the \textit{Fusion Expert}, each formed by coupling a specific prompt with the mask decoder. All experts share the same decoder architecture but maintain distinct parameters. During inference, all experts predict target-view masks, and the \textit{Post-hoc Cyclic Consistency Selector (PCCS)} adaptively selects the most reliable output in a non-parametric manner by evaluating the cross-view mask consistency.

\begin{figure}[t]
    \centering
    \includegraphics[width=0.98\linewidth]{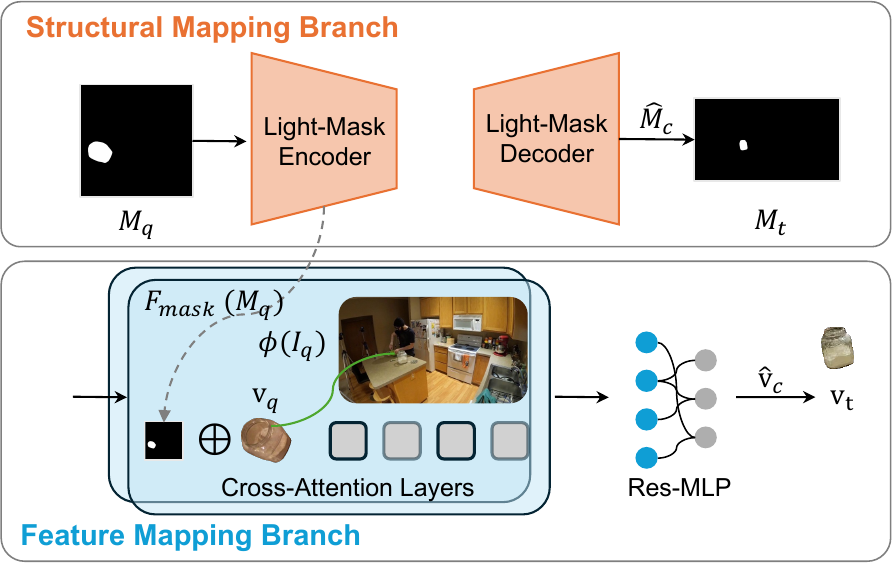}
    \caption{The structure of Visual Prompt Matcher. The Structural Mapping Branch is built upon a lightweight CNN-based mask encoder and decoder. The Feature Mapping Branch leverages Transformer-based cross-attention layers, while the Res-MLP component serves as a residual multi-layer perceptron.}
    \label{fig:vpfeature}
\end{figure}

\subsection{Cross-View Anchor Prompt Generator}
The V$^{2}$-Anchor module is primarily responsible for locating the corresponding object(s) in the target view $I_t$ by exploring geometric relationships between the query view $I_q$ and target view $I_t$. To achieve this, we leverage DINOv3, which encodes fine-grained, spatially aware representations at the patch level. Specifically, we extract patch-level features 
$\varphi(I_q)\!\in\!\mathbb{R}^{F\times H_q\times W_q}$ and 
$\varphi(I_t)\!\in\!\mathbb{R}^{F\times H_t\times W_t}$, 
and then perform a sequence of operations, including \textit{feature matching}, \textit{point stratification}, \textit{coordinate transformation}, and \textit{prompt encoding}, to generate geometry-based anchor prompts for cross-view segmentation.

\noindent \textbf{Feature Matching.}
For each query patch in $\varphi(I_q)$, we compute its similarity to all target patches $\varphi(I_t)$ to obtain a dense correspondence heatmap $\mathbf{H}$:
\begin{equation}
\mathbf{H}_{ij} =
\frac{
    \varphi(I_q)_i^{\top}\,\varphi(I_t)_j
}{
    \|\varphi(I_q)_i\|_2\,\|\varphi(I_t)_j\|_2
},
\label{eq:heatmap}
\end{equation}
where $\mathbf{H}_{ij}$ measures the cosine similarity between the $i$-th query patch and the $j$-th target patch, and $\mathbf{H}\in \mathbb{R}^{(H_qW_q)\times(H_tW_t)}$.
For each query patch, its most similar target patch is determined by $j^{*} = \arg\max_{j}\, \mathbf{H}_{ij}$
% \begin{equation}
% j^{*} = \arg\max_{j}\, \mathbf{H}_{ij},
% \end{equation}\label{j*}
and the corresponding 2D coordinates are recovered by mapping the linear indices of $i$ and $j^{*}$ back to their spatial patch locations in $I_q$ and $I_t$, respectively.

To restrict correspondence to the object region, the query mask $M_q$ is projected onto the DINOv3 patch grid to select foreground patches.  
% Each valid patch $i$ within $M_q$ forms the index set 
% $\mathcal{S}_q=\{\,i\mid(x_i,y_i)\!\in\!M_q\,\}$, 
Each valid patch $i$ within $M_q$ forms the point set $\mathcal{P}_q=\{\,i\mid(x_i,y_i)\!\in\!M_q\,\}$, 
and its best matched target point set $\mathcal{P}_t$ via feature matching.
% its best target match $j^{*}=\arg\max_j\mathbf{H}_{ij}$ is retained.  
This foreground-constrained matching suppresses background noise 
and concentrates on the object region, 
enhancing geometric precision for subsequent stratification.

\noindent \textbf{Point Stratification.}
Although applying $M_q$ suppresses some noises, $\mathcal{P}_t$ may still be redundant or locally clustered. Thus, we further apply a stratified sampling process that enforces a minimum pairwise distance $\tau$ among $\mathcal{P}_q$ and $\mathcal{P}_t$:
\begin{equation}
\mathcal{P'}_t = \{\,p_i \mid \|p_i - p_j\|_2 > \tau, \forall j<i \,\},
\label{eq:geo_stratify}
\end{equation}
resulting in a compact yet spatially diverse set of high-quality correspondences $\mathcal{P'}_t$.  
% These stratified points preserve global object geometry while discarding noisy or redundant local matches. effectively converting dense patch-level alignment into sparse point-level anchors.

\noindent \textbf{Coordinate Transformation and Prompt Encoding.}
The obtained sparse correspondences $\mathcal{P'}_t$ are defined in the original image coordinate system. To maintain consistency with SAM2’s internal representation, 
they are linearly transformed to the canonical coordinate space using a deterministic geometric projection $\Pi(\cdot)$: $\widetilde{\mathcal{P}}_t = \Pi(\mathcal{P'}_t;\, (H_t^{\mathrm{orig}}, W_t^{\mathrm{orig}}))$.
% The normalized coordinates $\widetilde{\mathcal{P}}_t$ are then formatted as point prompts and passed into the SAM2 prompt encoder, resulting in the coordinate-based anchor prompt $P^{q2t}_{anchor}$ for the target view.  
Finally, these coordinates $\widetilde{\mathcal{P}}_t$ are fed into the prompt encoder, resulting in the coordinate-based anchor prompt $P^{q2t}_{anchor}$ for the target view.
% This transformation preserves geometric relationships while enabling compatibility with the pretrained SAM2 architecture.

\subsection{Cross-View Visual Prompt Generator}
In contrast to V$^{2}$-Anchor, which focuses on geometric correspondence, the V$^{2}$-Visual is designed to capture appearance-based cues for identifying visually similar objects across views. We use region-level representations to guide segmentation similar to text-guided SAM~\cite{yuan2025sa2va}, and further introduce VPMatcher to bridge the cross-view appearance gap explicitly. Specifically, we first apply the SAM2 encoder $\phi(\cdot)$ to extract visual feature maps $\phi(I_q)$ and $\phi(I_t)$ from the query and target views, respectively. During training, we perform mask pooling on these feature maps using the corresponding object masks 
$M_q$ and $M_t$ to obtain region-level features:
\begin{equation}
    \mathbf{v}_q = \operatorname{MaskP}(\phi(I_{q}), M_{q}), \mathbf{v}_t = \operatorname{MaskP}(\phi(I_{t}), M_{t}). 
\end{equation}

\noindent \textbf{Visual Prompt Matcher (VPMatcher).} 
The detailed illustration of VPMatcher is given in Fig.~\ref{fig:vpfeature}. It consists of two branches, a \textit{Feature Mapping Branch} and a \textit{Structural Mapping Branch}. The former learns semantically consistent representations across views, while the latter reconstructs object masks to maintain structural coherence~\cite{liu2023modeling}.

\textit{(a) Feature Mapping Branch.}  Formally, given the fused prompt embedding $\mathbf{p}_f = \mathbf{v}_q + F_{\text{mask}}(M_q)$ obtained from the lightweight CNN-based mask encoder $F_{\text{mask}}(\cdot)$, the feature mapping branch applies linear projections to form the query, key, and value spaces: $\mathbf{q}, \mathbf{k}, \mathbf{v} = W_q \mathbf{p}_f, W_k \phi(I_q), W_v \phi(I_q)$. The attention weights as: $\alpha = \operatorname{Softmax}\left(\frac{\mathbf{q}\mathbf{k}^T}{\sqrt{D_e}}\right)$,
% \begin{equation} 
% \alpha = \operatorname{Softmax}\left(\frac{\mathbf{q}\mathbf{k}^T}{\sqrt{D_e}}\right), 
% \end{equation}
where $D_e$ is the embedding dimension. These weights are modulated by a spatial gating function to suppress background noise. The attention-weighted features are refined via multiple Transformer-based cross-attention encoder layers and a residual multi-layer perceptron, yielding the predicted cross-view embedding $\hat{\mathbf{v}}_{c}$.

\textit{(b) Structural Mapping Branch.} In parallel, the structural mapping branch reconstructs the target mask from the coarse geometric prior~\cite{williams2019deep,martens2016geometric}. The input mask $M_q$ is downsampled by $F_{\text{prior}}(\cdot)$ to a latent representation $\mathbf{m}_{\text{prior}}$. The prompt embedding $\mathbf{v}_q$ is projected to modulation parameters $(\gamma, \beta)$ through a feature-wise linear modulation~\cite{brockschmidt2020gnn} layer $\Phi_{\text{cond}}(\cdot)$, injecting semantic conditioning into the mask space: 
\begin{equation}
\tilde{\mathbf{m}} = \mathbf{m}_{\text{prior}} \odot (1 + \tanh(\gamma)) + \beta + F_{\text{mask}}(M_q). 
\end{equation}
Finally, the decoder $F_{\text{dec}}(\cdot)$ progressively upsamples $\tilde{\mathbf{m}}$ to generate the predicted cross-view mask $\hat{M}_c$. 
Furthermore, we can obtain new region features—i.e., the final visual prompt based on the predicted spatial mask as $\mathbf{v}_{c'} = \operatorname{MaskP}(\phi(I_{q}), \hat{M}_{c})$. This leads to the appearance-guided prompt, defined as $P^{q2t}_{visual} = \operatorname{MLP}\big([\hat{\mathbf{v}}_{c}, \mathbf{v}_{c'}]\big)$. Here, the query and target prototypes are concatenated and projected through a lightweight MLP to form the final visual prompt.

\begin{table*}
    \centering
    \resizebox{0.85\textwidth}{!}{
    \begin{tabular}{l ccc ccc c cc} 
   \toprule
         & \multicolumn{3}{c}{\textbf{Ego2Exo}}&  \multicolumn{3}{c}{\textbf{Exo2Ego}} & & \multicolumn{2}{c}{\textbf{Num. Param. (M)}} \\ 
         \cmidrule(lr){2-4} \cmidrule(lr){5-7} \cmidrule(lr){9-10}
         \textbf{Method} & IoU$\uparrow$ &  Cont.A$\uparrow$ &  Loc.E$\downarrow$ & IoU$\uparrow$ & Cont.A$\uparrow$ &  Loc.E$\downarrow$ & Total-IoU$\uparrow$  & Total & Train\\ \hline  
         PSALM \cite{zhang2024psalm} (Zero-shot) &  7.4 & 0.121 & 0.266 & 2.1 & 0.058 & 0.294 & 4.8 & 1587.1& 0\\  
         CMX \cite{zhang2023cmx} & 6.8 & 0.137 & 0.110 & 12.0 & 0.177 & 0.166 & 9.4 & 138.0 & 17.3 \\ 
         XSegTx \cite{grauman2024ego} &  18.9&  0.386&  0.070&  27.1& 0.358 & 0.104 & 23.0 & 12.1& 3.6\\   
         XMem \cite{grauman2024ego} &  19.3&  0.262&  0.151&   16.6 & 0.240& 0.160 & 18.0 & 62.2& 62.2\\  
         XMem + XSegTx \cite{grauman2024ego} &  34.9 & 0.559 & 0.038& 25.0 &  0.237& 0.117 & 30.0 & 75.6& 67.1\\ 
         Ref-SAM* & 29.2 & 0.452 & 0.077 & 42.2 & 0.502 & 0.096 & 37.8 & 224.8 & 4.3 \\ 
         ObjectRelator~\cite{fu2024objectrelator} & 35.3 & 0.540& \cellcolor{blue!15}{0.036} & 40.3 & 0.500 & \cellcolor{blue!25}{\textbf{0.068}} & 37.8 & 1587.3 &1587.3 \\ 
         O-MaMa (k-NN) ~\cite{mur2025mama}& 31.9 & 0.414& 0.195 & 30.9 & 0.373 & 0.127 & 31.4 & 154.0 &0 \\ 
         O-MaMa \cite{mur2025mama} & 42.6 &0.590 &\cellcolor{blue!25}{\textbf{0.033}}& 44.1 & 0.524 & 0.082 & 43.4 & 165.6 & 11.6 \\  \hline
         $\clubsuit$ \textbf{V$^{2}$-SAM} (Single-Expert) & \cellcolor{blue!15}{44.5} & \cellcolor{blue!15}{0.607} &0.055& \cellcolor{blue!15}{47.3} & \cellcolor{blue!15}{0.552} & 0.092 & \cellcolor{blue!15}{45.9} & {531.3} & {7.6} \\ 
         $\clubsuit$ \textbf{V$^{2}$-SAM} (Muti-Experts) & \cellcolor{blue!25}{\textbf{46.3}} & \cellcolor{blue!25}{\textbf{0.616}} &0.056& \cellcolor{blue!25}{\textbf{49.6}} & \cellcolor{blue!25}{\textbf{0.575}} & \cellcolor{blue!15}{0.078} & \cellcolor{blue!25}{\textbf{48.0}} & {543.4} & {15.3} \\ 
         % \bottomrule
         \midrule
    \end{tabular}
    }
    \caption{Results on the Ego-Exo4D Correspondences v2 test split.}
    \label{tab:state_of_art}
\end{table*}

\subsection{Multi-Expert Training}
Intuitively and empirically, we find that different prompts specialize in distinct forms of reasoning: V$^2$-Anchor is grounded in geometric structure, while V$^2$-Visual focuses on visual appearance. To leverage their complementary strengths and handle diverse cross-view scenarios, where both localization and visual association are crucial, especially in challenging cases, we design a multi-expert training mechanism that adaptively selects the most suitable expert under varying scene conditions.

To that end, as introduced in Sec.~\ref{sec:overview}, three experts are constructed and trained, i.e., \textit{Anchor Expert}, \textit{Visual Expert}, and \textit{Fusion Expert}. More specifically, all of them are using the same mask decoder architecture but different prompts as input to condition the mask decoder: Anchor Expert is using the geometry-aware prompt $P^{q2t}_{anchor}$ generated from V$^2$-Anchor; Visual Expert is using visual-guided prompt $P^{q2t}_{visual}$ predicted from V$^2$-Visual, and Fusion Expert takes both $P^{q2t}_{anchor}$ and $P^{q2t}_{visual}$, and fuse them into a unified prompt embedding.
For training, note that all steps in V$^{2}$-Anchor are non-learnable and directly output coordinate prompts $P^{q2t}_{anchor}$ in the target view. Since these coordinate-based prompts are already supported by SAM2, we keep their corresponding expert entirely training-free by reusing the pretrained SAM2 mask decoder. In contrast, the Visual and Fusion Experts introduce the VPMatcher with new learnable parameters, and their resulting visual or visual–anchor prompts extend beyond the original SAM2 design. Therefore, we train the VPMatcher and mask decoder to better adapt to our task and datasets, while keeping the SAM2 encoder frozen. Both the Visual and Fusion Experts are optimized using the loss functions described below.

% \subsubsection{Loss Functions}
\noindent\textbf{Loss Functions.}
Our training objective consists of three components: a visual contrastive loss $\mathcal{L}_{v}$, 
a structural constraint loss $\mathcal{L}_{s}$, and a mask prediction loss $\mathcal{L}_{m}$. 
The total loss $\mathcal{L}$ is defined as a weighted combination of these terms:
\begin{equation}
\begin{aligned}
    \mathcal{L} &=\lambda_{1}\mathcal{L}_{v}\!\left(\hat{\mathbf{v}}_{c}, \mathbf{v}_{t}\right) 
     + \lambda_{2}\mathcal{L}_{s}\!\left(\hat{M}_c, M_t\right)  + \lambda_{3}\mathcal{L}_{m}\!\left(\hat{M}_t, M_t\right),
\end{aligned}
\end{equation}
where $\lambda_{1}$, $\lambda_{2}$, and $\lambda_{3}$ are the balancing coefficients.

\textit{(a) Visual Contrastive Loss.} We employ contrastive loss across different views to enforce cross-view region-level features mapping in the same setting as~\cite {pan2023prior,pan2024pir}. This loss encourages positive pairs to be close together and negative pairs to be far apart in the embedding space, thereby forming compact, well-separated feature clusters. Formally, it is defined as:
\begin{equation}
\begin{alignedat}{2}
\mathcal{L}_v
= -\frac{1}{N} \sum_{i=1}^{N} \Bigg[
& \log \frac{
    \exp\!\left(\mathrm{sim}(\mathbf{v}_c,\mathbf{v}_t)/\tau\right)
}{
    \sum_{k=1}^{N} \exp\!\left(\mathrm{sim}(\mathbf{v}_c,\mathbf{v}_t^k)/\tau\right)
}
\\[-3pt]
& \quad +
\log \frac{
    \exp\!\left(\mathrm{sim}(\mathbf{v}_t,\mathbf{v}_c)/\tau\right)
}{
    \sum_{k=1}^{N} \exp\!\left(\mathrm{sim}(\mathbf{v}_t,\mathbf{v}_t^k)/\tau\right)
}
\Bigg].
\end{alignedat}
\end{equation}
where $\mathrm{sim}(\cdot)$ denotes the cosine similarity function and $\tau$ is a temperature parameter.

% \textit{Structural Constraint and Mask Loss.}
\textit{(b) Mask Prediction Loss.}
Both the spatial constraint loss $\mathcal{L}_{s}$ and the mask loss $\mathcal{L}_{m}$ 
are designed based on mask-level supervision to ensure accurate spatial localization and structural consistency. The mask loss combines a pixel-wise cross-entropy term~\cite{mao2023cross} and a region-level Dice loss~\cite{sudre2017generalised}:
\begin{equation}
    \mathcal{L}_{m} =
    \mathcal{L}_{CE}(\hat{M}, M) +
    \mathcal{L}_{Dice}(\hat{M}, M),
\end{equation}
where $\mathcal{L}_{CE}$ penalizes pixel-wise classification errors, 
and $\mathcal{L}_{Dice}$ measures the overlap between the predicted and ground-truth masks.

\textit{(c) Structural Constraint Loss.}
Specifically, we also applied the mask loss for constraint cross-view structural mapping. $\mathcal{L}_{s}$ imposes a constraint on the cross-view structural mapping in VPMatcher, encouraging the model to learn a fixed spatial transformation.

\begin{table}[t]
    \centering
    \resizebox{0.84\columnwidth}{!}{%
    \begin{tabular}{l c c c}
    \toprule
        \textbf{Method} & \makecell{$\mathcal{J}\&\mathcal{F}_{m}$\\$\uparrow$} & \makecell{$\mathcal{J}_{m}$\\$\uparrow$} & \makecell{$\mathcal{F}_{m}$\\$\uparrow$} \\ \hline
        SiamMAE~\cite{NEURIPS2023_7ffb9f1b}  & 60.7 & 58.4 & 62.9 \\
        CrocoV2 + Cont. Pretrain~\cite{Weinzaepfel_2023_ICCV}  & 40.0 & 37.4 & 42.5 \\
        Probabilistic Warp Consistency~\cite{Truong_2022_CVPR}  & 42.9 & 42.6 & 42.7 \\
        DINO ViTs/8~\cite{Caron_2021_ICCV}  & 64.5 & 61.6 & 67.5 \\
        DINO ViTb/8~\cite{Caron_2021_ICCV}  & 66.4 & 63.7 & 69.2 \\
        DINOv2 + Reg ViTb/14~\cite{darcet2024visiontransformersneedregisters} & 62.1 & 59.6 & 64.8 \\ 
        PCC~\cite{baade2025self} & \cellcolor{blue!15}{70.2} & \cellcolor{blue!15}{67.8} & \cellcolor{blue!15}{72.7} \\ \hline
        $\clubsuit$ \textbf{V$^{2}$-SAM (Ours)} & \cellcolor{blue!25}{\textbf{78.8}} & \cellcolor{blue!25}{\textbf{76.5}} & \cellcolor{blue!25}{\textbf{81.0}} \\ 
        % \bottomrule
        \midrule
    \end{tabular}
    }
    \caption{Comparison of video object correspondence on DAVIS-2017 Val with a temporal gap of 20 frames.}
    \label{tab:pcc_results}
\end{table}

\subsection{Post-hoc Cyclic Consistency Selector}
Our PCCS is designed to dynamically identify the most reliable expert, based on the bidirectional consistency between the query and target masks.
% Given the predicted target masks from all three experts, PCCS maps them back to the query view and measures its geometric alignment with the original prompt mask $M_{q}$. This is achieved by feeding each prediction $\hat{M}_{t}^{k}$, where $k$ means index of expert, to the V$^{2}$-Anchor. 
% Notably, the mapping direction is intentionally reverse, using $I_t$ as the query and $I_q$ as the target, so that the target-view prediction is projected back to the query frame.
PCCS projects each expert’s prediction $\hat{M}_{t_{k}}$ back to the query view via the V$^{2}$-Anchor to assess geometric consistency with $M_q$.
Formally, this inverse correspondence process can be expressed as:
\begin{equation}
P^{t2q}_
{k} = \operatorname{V^{2}\!-\!Anchor}\left(I_{t}, I_{q}; \hat{M}_{t_{k}}\right),
\label{eq:sparsecorr}
\end{equation}
where $\hat{M}_{t_{k}}$ denotes the prediction from the $k$-th expert, and $P^{t2q}_{k}$ represents the back-projected points on the query image corresponding to pixels within that prediction.

For cyclic consistency validation, we compute the average distance between the back-projected points and reference points sampled from the query mask as a proxy score to select the expert.
Unlike prior cyclic schemes that reconstruct a query-view mask, our PCCS operates directly at the point level, avoiding an extra decoding pass while preserving selection accuracy. This lightweight validation enables effective expert selection during inference.

% \begin{align}
% d_{k} &= \frac{1}{\left|P_{\mathrm{ref}}^{q}\right|} 
% \sum_{i}\left\|\hat{P}_{k, i}^{q}-P_{\mathrm{ref}, i}^{q}\right\|_{2} \label{eq:dk}\\
% M^{*} &= \hat{M}_{t_{k^{*}}}, \quad 
% k^{*} = \arg \min_{k} d_{k} \label{eq:select_k}
% \end{align}

% Unlike prior cyclic schemes~\cite{} that reconstruct the query-view mask from projected points (Cycle-Mask), our Cycle-Points design operates directly at the geometric correspondence level. This avoids passing the points back through the decoder, significantly reducing computation while achieving comparable selection accuracy. Empirically, experts whose masks yield geometrically coherent back-projections (i.e., smaller $d_k$) are also those that achieve higher IoU consistency across views, validating the point-level cyclic distance as a reliable proxy for cross-view correctness.

% For cyclic validation, we compare these predicted points with the reference points \todo{$P^{q}_{\text{ref}}$ sampled from the raw prompt mask in $I_q$.} The average point-wise Euclidean distance serves as the cyclic consistency score for the $k$-th expert, and the expert with the smallest distance is selected as the final output:

%% file: sec/4_results.tex
\section{Experiments}
\label{sec:results}

% \begin{table}[t]
%     \centering
%     \resizebox{\columnwidth}{!}{%
%     \begin{tabular}{l l l c}
%         \textbf{Setting} & \textbf{Method} & \textbf{Dataset} & IoU$\uparrow$ \\ \midrule
%         \multirow{6}{*}{ZSL} 
%             & XSegTx~\cite{grauman2024ego} & COCO / Self-Gen. Pairs & 1.5 \\
%             & SEEM~\cite{NEURIPS2023_3ef61f7e} & COCO Panoptic, RefCOCO/+g & 2.5 \\
%             & PSALM~\cite{zhang2024psalmpixelwisesegmentationlarge} & COCO Panoptic, RefCOCO/+g, etc. & 14.2 \\
%             & PSALM~\cite{zhang2024psalmpixelwisesegmentationlarge} & Ego-Exo4D & 39.9 \\
%             & ObjectRelator~\cite{fu2024objectrelator} & Ego-Exo4D & 42.8 \\
%             \midrule
%             & \textbf{V$^{2}$-SAM} (Single-Expert) & Ego-Exo4D & 66.4 \\
%             & \textbf{V$^{2}$-SAM} (Multi-Experts) & Ego-Exo4D & \textbf{77.2} \\ \midrule
%         % \multirow{3}{*}{Retrained}
%         %     & PSALM & HANDAL-X & 83.4 \\
%         %     & ObjectRelator & HANDAL-X & 84.7 \\
%         %     & V$^{2}$-SAM (Single-expert) & HANDAL-X & -- \\
%         %     & \textbf{V$^{2}$-SAM (Multi-Experts)} & HANDAL-X & \textbf{--} \\ \bottomrule
%     \end{tabular}
%     }
%     % \caption{Results on the \textsc{HANDAL-X} dataset. Comparison of zero-shot (ZSL) and retrained models on HANDAL-X and related datasets.}
%     \caption{Comparison of zero-shot (ZSL) object segmentation on HANDAL-X.}
%     \label{tab:handalx_iou}
% \end{table}

\begin{table}[t]
    \centering
    \resizebox{\columnwidth}{!}{%
    \begin{tabular}{l l l c}
    \toprule
        \textbf{Setup} & \textbf{Method} & \textbf{Dataset} & \textbf{IoU}$\uparrow$ \\ \midrule
        \multirow{7}{*}{ZSL} 
            & XSegTx~\cite{grauman2024ego} & COCO / Self-Gen. Pairs & 1.5 \\
            & SEEM~\cite{NEURIPS2023_3ef61f7e} & COCO Panoptic, RefCOCO/+g & 2.5 \\
            & PSALM~\cite{zhang2024psalmpixelwisesegmentationlarge} & COCO Panoptic, RefCOCO/+g, etc. & 14.2 \\
            & PSALM~\cite{zhang2024psalmpixelwisesegmentationlarge} & Ego-Exo4D & 39.9 \\
            & ObjectRelator~\cite{fu2024objectrelator} & Ego-Exo4D & 42.8 \\
            \cmidrule(lr){2-4}
            & $\clubsuit$ \textbf{V$^{2}$-SAM} (Single-Expert) & Ego-Exo4D & \cellcolor{blue!15}{66.4} \\
            & $\clubsuit$ \textbf{V$^{2}$-SAM} (Multi-Experts) & Ego-Exo4D & \cellcolor{blue!25}{\textbf{77.2}} \\ 
            \midrule
            % \bottomrule
    \end{tabular}
    }
    \caption{Comparison of zero-shot (ZSL) object segmentation on HANDAL-X.}
    \label{tab:handalx_iou}
\end{table}

\subsection{Experimental Setup}
\noindent\textbf{Datasets and Evaluation Metrics.}
We conduct experiments on three representative benchmarks: 
Ego-Exo4D~\cite{grauman2024ego} for ego–exo object correspondence, DAVIS-17~\cite{pont20172017} for video object tracking, and HANDAL-X~\cite{guo2023handal} for robotic-ready cross-view segmentation. Ego-Exo4D is evaluated using \textit{Mean Intersection over Union (mIoU)}, \textit{Continuity Accuracy (Cont.A)}, and \textit{Localization Error (Loc.E)}. 
For DAVIS-17, we report \textit{Region Similarity} ($\mathcal{J}\!\uparrow$) and \textit{Contour Accuracy} ($\mathcal{F}\!\uparrow$). HANDAL-X is evaluated with mIoU.

\noindent \textbf{Implementation Details.} We employ the SAM2-Hiera-Large model as the base segmentation backbone, and a DINOv3 ViT-L/16 Encoder as the feature extractor. All experiments were completed on eight NVIDIA H-series GPUs. The batch size for each GPU is set to 16, and with $4 \times 10^{-5}$ learning rate. The loss weights are set to $\lambda_{1} : \lambda_{2} : \lambda_{3} = 1:1:10$. However, during the first 4K training steps, we adopt a contrastive learning objective with $\lambda_{1} = 100$ to enhance feature representation. 
During inference, all three experts operate in parallel, generating candidate masks conditioned on the visual and geometric anchor prompts. The final prediction is determined by our PCCS, with all evaluations performed using a batch size of 1.

\subsection{Main Results}
\noindent \textbf{Baselines and Competitors.}
We compare V$^{2}$-SAM with representative baselines on the three correspondence benchmark. 
XSegTx and XMem~\cite{grauman2024ego} are the official baselines that extend co-segmentation and temporal memory models for cross-view propagation. 
CMX~\cite{zhang2023cmx} fuses multi-modal inputs through a transformer-based decoder. 
PSALM~\cite{zhang2024psalm} integrates a large language model with Mask2Former~\cite{mask2former} for zero-shot segmentation, 
while ObjectRelator~\cite{fu2024objectrelator} enhances PSALM with cross-view relational reasoning. Ref-SAM* is a SAM2-adapted baseline via visual reference-guided prompts. The recent O-MaMa~\cite{mur2025mama} treats cross-view segmentation as a mask-matching task by generating FastSAM candidates and selecting the best match via contrastive learning. 
All methods are evaluated on the Ego-Exo4D Correspondences v2 test split.

\noindent \textbf{Results on Ego-Exo4D.}
The comparison results are summarized in Tab.~\ref{tab:state_of_art}. 
We report IoU$\uparrow$, Cont.A$\uparrow$, and Loc.E$\downarrow$ for both \textit{Ego2Exo} and \textit{Exo2Ego} directions, 
along with the averaged total IoU and parameter statistics.  
Among existing baselines, O-MaMa achieves the highest IoU of 42.6/44.1 on the two directions, 
while ObjectRelator attains 35.3/40.3 with substantially larger model size (1.6B parameters). 
Our proposed V$^{2}$-SAM (Multi-Experts) establishes a new state-of-the-art, 
achieving 46.31 IoU on Ego2Exo and 49.61 on Exo2Ego, 
surpassing O-MaMa by 3.7 and 5.5 IoU points on the Ego2Exo and Exo2Ego directions, respectively. 
Even the single-expert version already surpasses all prior baselines, 
achieving a total IoU of 45.9, which is 2.5 points higher than O-MaMa (43.4 IoU), demonstrating the strong capacity of our model.  Compared to the Ref-SAM*, visual-reference guided segmentation struggles with cross-view segmentation, particularly in the Ego2Exo task, indicating that our framework addresses the shortcomings in visual guidance and achieves further improvements on this basis.
Notably, our multi-expert model achieves this performance with only 543M total parameters and 15M trainable parameters, 
which is $\sim$1\% of ObjectRelator’s parameter count yet yields a +10.2 total IoU improvement.  
These results validate the effectiveness of our Multi-Prompt Experts.

\begin{table}[t]
    \centering
    \resizebox{0.78\columnwidth}{!}{%
    \begin{tabular}{lccc}
     \toprule
        \textbf{Variant} & \makecell{\textbf{Ego2Exo}\\IoU$\uparrow$} & \makecell{\textbf{Exo2Ego}\\IoU$\uparrow$} & \makecell{\textbf{Total}\\IoU$\uparrow$} \\ 
        \midrule
        \multicolumn{4}{l}{\textit{Anchor Expert}} \\ 
        \midrule
        w/o V$^{2}$-Anchor  & 0.8 & 2.1 & 1.5 \\
        \textbf{w/ V$^{2}$-Anchor (Ours)} 
            & \cellcolor{blue!25}{\textbf{38.7}} 
            & \cellcolor{blue!15}{\textbf{41.6}} 
            & \cellcolor{blue!15}{\textbf{40.1}} \\ 
        \midrule
        \multicolumn{4}{l}{\textit{Visual Expert}} \\ 
        \midrule
        w/o V$^{2}$-Visual & 1.0 & 5.0 & 3.0 \\
        \textbf{w/ V$^{2}$-Visual (Ours)} 
            & \cellcolor{blue!15}{\textbf{36.2}} 
            & \cellcolor{blue!25}{\textbf{46.6}} 
            & \cellcolor{blue!25}{\textbf{41.4}} \\ 
        % \bottomrule
        \midrule
    \end{tabular}
    }
    \caption{Ablation on individual experts.}
    \label{tab:ablation_experts_single}
\end{table}

% \wrz{\subsection{Main Results on DAVIS}
\noindent \textbf{Results on DAVIS-17.}
To further assess the generality of our framework for cross-view correspondence, we evaluate it on DAVIS-17, a standard benchmark for video object tracking.
Following~\cite{baade2025self}, we evaluate correspondence accuracy on all DAVIS-17 frame pairs separated by 20 frames, and consider only objects that are simultaneously visible in both views. We report performance using the standard $\mathcal{J}\&\mathcal{F}_{m}$, $\mathcal{J}_{m}$, and $\mathcal{F}_{m}$ metrics. 
As shown in Tab.~\ref{tab:pcc_results}, our V$^{2}$-SAM achieves the highest performance on DAVIS-17 across all evaluation metrics, 
surpassing all existing self-supervised correspondence models by a clear margin. 
V$^{2}$-SAM also achieves strong correspondence quality across large temporal gaps,
highlighting its potential for robust object tracking under significant viewpoint shifts.
% Compared with the challenging viewpoint changes in Ego–Exo4D, DAVIS is relatively simpler yet offers diverse scenes and realistic object motions, making it a valuable testbed for common scenarios.
% We compare against representative self-supervised correspondence baselines employed in prior PCC work, including SiamMAE~\cite{pcc}, CroCo v2~\cite{pcc}, DINO variants~\cite{pcc}, and the iterative PCC refinement.
% The dataset contains in-the-wild videos with unconstrained camera motion and diverse scene conditions, providing a strong testbed for examining cross-view robustness beyond the Ego–Exo4D domain.

% \wrz{\subsection{Main Results on HANDAL-X}
\noindent \textbf{Results on HANDAL-X.}
HANDAL-X contains real-world, robotics-ready manipulable objects, offering a more practical testbed for Human2Robot transfer than Ego–Exo4D.  
Following~\cite{fu2024objectrelator}, we evaluate V$^{2}$-SAM in the ZSL setting using a model trained on Ego–Exo4D. Tab.~\ref{tab:handalx_iou} shows that V$^{2}$-SAM demonstrates strong cross-view generalization, 
even the single-expert variant reaches an IoU of 66.4, substantially outperforming PSALM (39.9 IoU) and ObjectRelator (42.8 IoU). The multi-experts version further increases performance to 77.2 IoU. 
% We further retrain our model on HANDAL-X. As shown in Tab.~\ref{tab:handalx_iou}, V$^{2}$-SAM also achieves state-of-the-art performance in this setting, confirming that the proposed architecture exhibits strong cross-domain adaptability.
% we evaluate the potential of V$^{2}$-SAM on HANDAL-X within the context of Embodied AI.

% \begin{table}[t]
%     \centering
%     \resizebox{0.8\columnwidth}{!}{%
%     \begin{tabular}{lccc}
%         % \toprule
%         \textbf{Method} & \makecell{Ego2Exo\\IoU$\uparrow$} & \makecell{Exo2Ego\\IoU$\uparrow$} & \makecell{Total\\IoU$\uparrow$} \\ 
%         \midrule
%         A: Anchor Expert & {38.7} & {41.6} & {40.1} \\
%         B: Visual Expert & {36.2} & {46.6} & {41.4} \\
%         C:  Fusion Expert & {44.5} & {47.3} & {45.9} \\ 
%         \midrule
%         A + B & {42.7} & {48.2} & {45.5} \\ 
%         A + B + C & {\textbf{46.3}} & {\textbf{49.6}} & {\textbf{48.0}} \\ 
%         \bottomrule
%     \end{tabular}
%     }
%     \caption{Results of different expert decoder combinations on v2 test split.
%     A: Anchor Expert, 
%     B: Visual Expert, 
%     C: Fusion Expert.}
%     \label{tab:sota_val_v1}
% \end{table}

\begin{table}[t]
    \centering
    \resizebox{0.76\columnwidth}{!}{%
    \begin{tabular}{lccc}
    \toprule
        \textbf{Method} & \makecell{\textbf{Ego2Exo}\\IoU$\uparrow$} & \makecell{\textbf{Exo2Ego}\\IoU$\uparrow$} & \makecell{\textbf{Total}\\IoU$\uparrow$} \\ 
        \midrule
        A: Anchor Expert & 38.7 & 41.6 & 40.1 \\
        B: Visual Expert & 36.2 & 46.6 & 41.4 \\
        C: Fusion Expert & \cellcolor{blue!15}{44.5} & 47.3 & \cellcolor{blue!15}{45.9} \\ 
        \midrule
        A + B & 42.7 & \cellcolor{blue!15}{48.2} & 45.5 \\ 
        A + B + C & \cellcolor{blue!25}{\textbf{46.3}} & \cellcolor{blue!25}{\textbf{49.6}} & \cellcolor{blue!25}{\textbf{48.0}} \\ 
        \midrule
        % \bottomrule
    \end{tabular}
    }
    \caption{Results of different expert decoder combinations on v2 test split.
    A: Anchor Expert, 
    B: Visual Expert, 
    C: Fusion Expert.}
    \label{tab:sota_val_v1}
\end{table}

\subsection{Ablation Study}

\noindent \textbf{Ablation on Individual Expert.}
To evaluate the contribution of each expert, we compare it against its simplified baseline.
1) For the \textit{Anchor Expert}, we replace V$^{2}$-Anchor with a simplified baseline that directly uses the centroid of $M_{q}$ as the prompt for target-view prediction. 
For the \textit{Visual Expert}, V$^{2}$-Visual aligns ego–exo object representations from both feature and structural perspectives. 
To evaluate its effectiveness, we remove the Visual Prompt Matcher and directly use the mask-pooled query-view feature as the prompt for cross-view segmentation.
Tab.~\ref{tab:ablation_experts_single} summarizes the results. 
For Anchor Expert, removing V$^{2}$-Anchor leads to near-zero performance (1.5 total IoU), as the model entirely loses the ability to localize geometric anchors across views.
For Visual Expert, removing the V$^{2}$-Visual also causes drastic degradation (3.0 total IoU), as directly using query-view features cannot provide meaningful cues for the target view. In contrast, our Visual Expert with V$^{2}$-Visual achieves strong performance (41.4 total IoU), 
effectively bridging the cross-view appearance gap.

% This also reflects a key factor limiting vanilla SAM's ability to achieve cross-view localization.
% V$^{2}$-Anchor enables cross-view object localization by deriving reliable geometric anchors in the target view.

% 画一个比较fancy的图：decoder场景分布分析。最好画一个雷达图，信息量要大，不能仅仅是iou。比如不同iou差异阈值筛选出来的数据，最后在场景分析上得到的结论是一致的，都是sparse在cooking上更强
\begin{figure}
    \centering
    \includegraphics[width=0.99\linewidth]{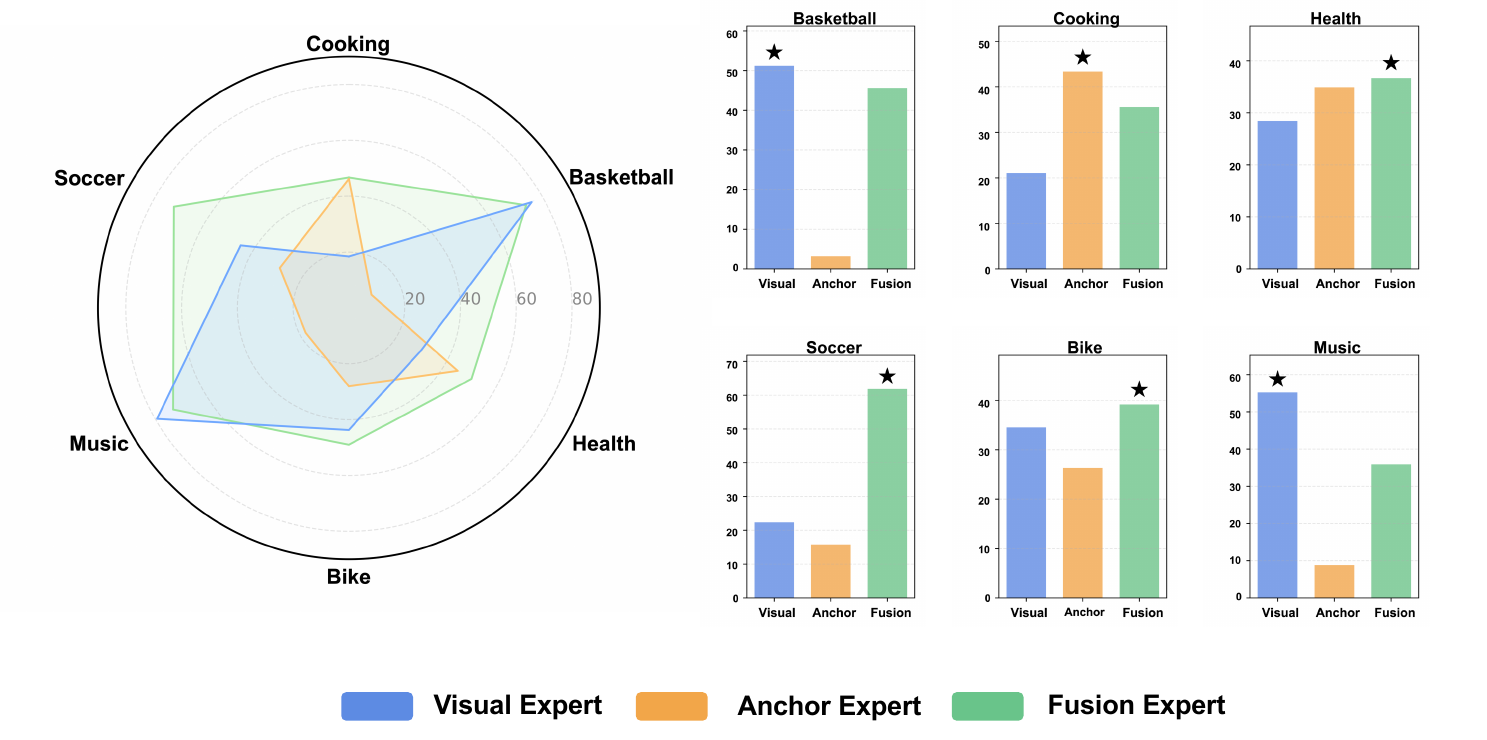}
    \caption{Comparison of Anchor, Visual, and Fusion Experts across different scenes. Left: per-scene IoU radar plot for the three experts.
Right: per-scene Win\% bars showing PCCS selections.}
    \label{tab:expert-performance}
\end{figure}

\noindent \textbf{Ablation on Experts Combinations.}
As shown in Tab.~\ref{tab:sota_val_v1}, due to their single-prompt nature, neither the Anchor Expert nor the Visual Expert achieves strong performance on Ego2Exo or Exo2Ego when evaluated individually.
Integrating the two experts produces a notable performance gain (Ego2Exo: 38.7/36.2 → 42.7), confirming that anchor and visual cues provide complementary cross-view information. 

The Fusion Expert, which reasons jointly over both anchor and visual prompts, achieves balanced performance and robustly handles diverse scene conditions, 
resulting in performance that is on par with the two-expert combination.
Finally, integrating all three experts within our Multi-Prompt Experts framework produces the best overall results across both Ego2Exo and Exo2Ego directions.
% To fully exploit the complementary strengths of geometric and visual prompts, we adopt a multi-expert design that adaptively selects the most reliable expert for each instance.
% This confirms that integrating multiple experts with PCCS provides a unified mechanism for handling diverse cross-view scenarios.

% \noindent \wrz{\textbf{Ablation on the Cyclic Selector.}
% Tab. \ref{tab:cycle_selector_ablation} compares our Cycle-Points Selector with the prior Cycle-Mask Selector, which reconstructs query-view masks for cyclic validation.
% Instead of predicting masks, our method directly measures geometric consistency between predicted query points and sampled reference points from the raw prompt mask, selecting the expert with the smallest cyclic distance.
% Despite its simpler design, Cycle-Points achieves comparable or even higher accuracy while significantly reducing computational cost.
% For the two-expert setup (A+B), it improves Exo→Ego IoU by +1.4 and shortens runtime by 110 ms per sample.
% With three experts (A+B+C), it maintains accuracy while reducing both latency and FLOPs.

% These results demonstrate that expert selection can be effectively driven by point-level cyclic consistency alone, providing a lightweight yet equally reliable alternative to mask-based cyclic reasoning.
% }

% \noindent \wrz{\textbf{Ablation on the Proposed modules.}}

\noindent \textbf{Expert Performance Analysis.}
To better understand how each expert behaves across different scene types, Fig.~\ref{tab:expert-performance} visualizes their performance on the Ego2Exo task.
Results are computed on samples where the IoU difference exceeds 0.3, highlighting cases with substantial disagreement and allowing clearer analysis of expert specialization. We observe that the 1) Anchor Expert demonstrates strong performance in structured and relatively static environments such as Cooking and Health. However, its effectiveness diminishes in scenes involving large motion or significant appearance change.
In contrast, 2) the Visual Expert excels in dynamic and human-centric scenarios such as Basketball and Music, where appearance cues, semantic consistency, and deformation tolerance are crucial. 3) The Fusion Expert shows consistently broad coverage across all scene categories in the radar plot, indicating stable and balanced performance across diverse scenes. 

\subsection{Qualitative Results}
To provide a more intuitive understanding of V$^{2}$-SAM, we present qualitative examples for both the Ego2Exo (\cref{fig:qualitative1}) and Exo2Ego (\cref{fig:qualitative2}) tasks. For each sample, we visualize the predictions from the \textit{Anchor Expert} (AE), \textit{Visual Expert} (VE), and \textit{Fusion Expert} (FE), getting the final mask selected by PCCS. 1) \textit{Ego2Exo:} The VE may incorrectly segment visually similar objects in cluttered backgrounds with multiple distractors, whereas the AE can accurately localize the correct target guided by geometric cues. Conversely, in highly dynamic scenes such as basketball, the VE remains reliable. 2) \textit{Exo2Ego:} object manipulation scenes often involve substantial hand occlusion in the ego view requires both geometric anchoring to locate the object and visual cues to determine its boundaries. The FE performs well under these conditions. For scenes where object appearance changes dramatically across views, such as music, the AE struggles to maintain consistent geometric correspondence, whereas the VE remains robust to such variations.

From the visualization results, we also observe that the Fusion Expert exhibits stable performance across diverse scene conditions, benefiting from its use of complementary prompts. PCCS further exploits cross-view mask consistency to reliably select the best prediction for each instance.

%% file: sec/5_conclusions.tex
\section{Conclusions}
\label{sec:conclusion}

\begin{figure}
    \centering
    \includegraphics[width=0.99\linewidth]{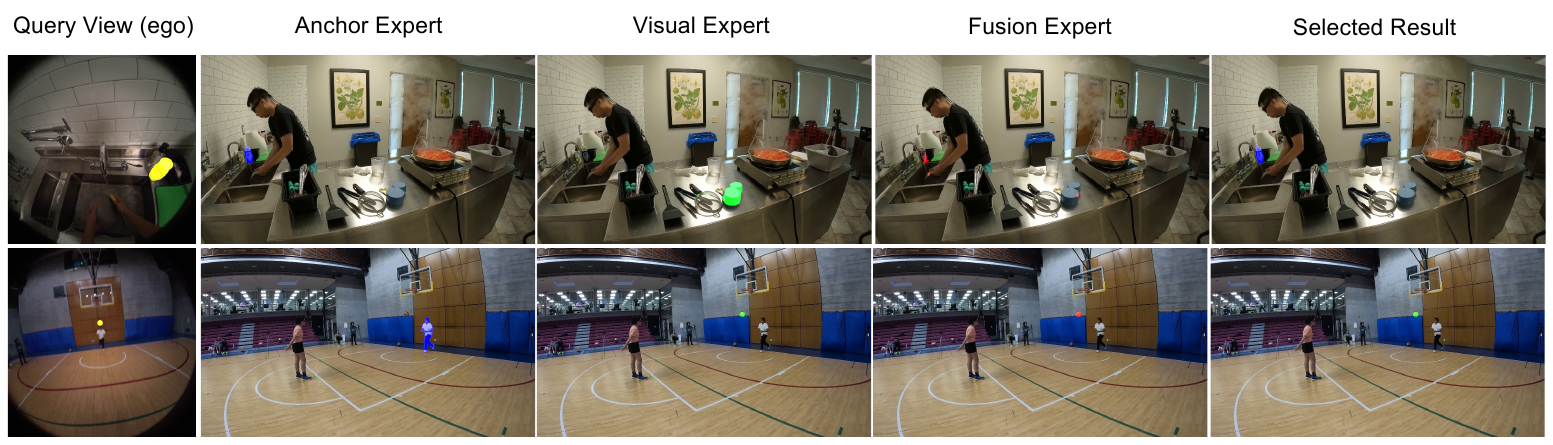}
    \caption{Ego2Exo qualitative results. From left to right: query view, predictions from the Anchor Expert, Visual Expert, and Fusion Expert, followed by the final output selected by the PCCS.}
    \label{fig:qualitative1}
\end{figure}

\begin{figure}
    \centering
    \includegraphics[width=0.95\linewidth]{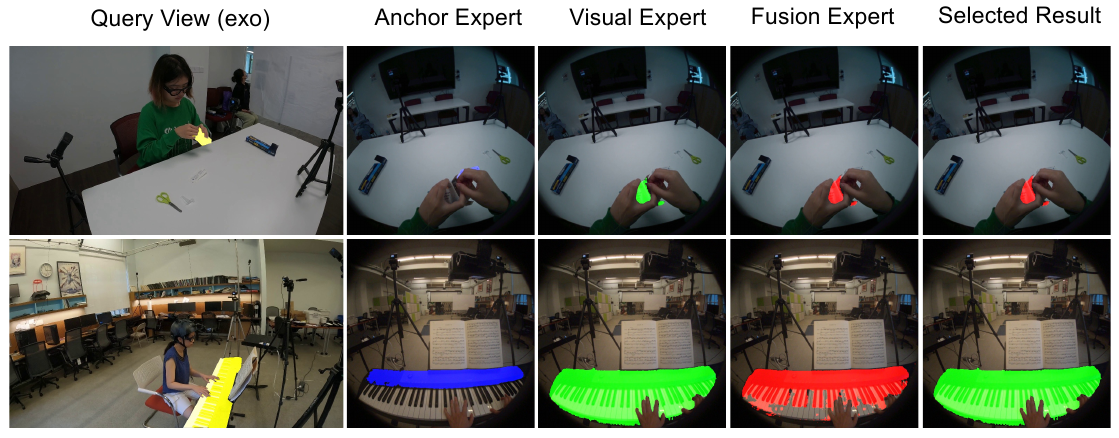}
    \caption{Exo2Ego qualitative results. From left to right: query view, predictions from the Anchor Expert, Visual Expert, and Fusion Expert, followed by the final output selected by the PCCS.}
    \label{fig:qualitative2}
\end{figure}

In this work, we presented V$^{2}$-SAM, a unified framework that extends the powerful segmentation foundation model SAM2 to the challenging task of cross-view object correspondence, with a particular focus on the Ego–Exo setting. To overcome the inherent difficulties posed by drastic viewpoint and appearance variations, we introduced two complementary prompt generators: V$^{2}$-Anchor, which restores spatially grounded prompting through geometry-aware correspondence, and V$^{2}$-Visual, which enhances appearance-guided understanding via a novel visual prompt matcher. Building on these, we designed a multi-prompt expert framework and a Post-hoc Cyclic Consistency Selector that adaptively selects the optimal expert for each instance. Extensive experiments on Ego-Exo4D, DAVIS-2017, and HANDAL-X benchmarks demonstrate that V$^{2}$-SAM achieves state-of-the-art performance across diverse cross-view correspondence tasks.

% We hope our study not only establishes a strong baseline for cross-view segmentation but also inspires future exploration of integrating spatial and visual prompting for broader multi-view and embodied perception scenarios.

%% file: sec/7_suppl.tex
\clearpage
\maketitlesupplementary

% 从这里开始章节编号从 1 开始
\setcounter{section}{0}
\setcounter{subsection}{0}

\tableofcontents
\addtocontents{toc}{\protect\setcounter{tocdepth}{2}}
\section{More Related Work}

\subsection{Segment Anything Model}
The Segment Anything Model (SAM) is a prompt-driven foundation model for universal image localization~\cite{pan2025enhance,liu2025control,liu2025diverse} and segmentation, capable of producing high-quality masks from simple inputs like points or bounding boxes. It has inspired domain-specific extensions such as MedSAM~\cite{zhou2025medsam} for medical imaging, InstructSAM~\cite{zheng2025instructsam} for remote sensing, and video or language-aware variants that enhance temporal and semantic reasoning across diverse tasks. 

SAM2~\cite{ravi2024sam2segmentimages} is one of the most powerful segmentation models to date, demonstrating outstanding generalization ability in images and videos. It consists of a \textit{Image Encoder} for visual feature extraction, a \textit{Prompt Encoder} for embedding points, boxes, or masks, and a \textit{Mask Decoder} that predicts object masks using features from both encoders. A memory mechanism further enables temporal mask propagation across video frames. However, since all prompts are defined within the coordinate system of the target image, SAM2 is inherently non-trivial to adapt to cross-view scenarios, where object position, scale, shape, and appearance often vary drastically across views.

Our framework is built upon SAM2, retaining its core components. Specifically, we keep the \textit{SAM2 Encoder} $\phi(\cdot)$, \textit{Prompt Encoder}, and \textit{Mask Decoder}, while discarding memory-related modules to focus on frame-level correspondence. This design allows the framework to generalize seamlessly across both cross-view image and video tasks.  
To unlock SAM2’s potential for cross-view object correspondence, we introduce four novel modules:  
1) a \textit{Cross-View Anchor Prompt Generator (V$^{2}$-Anchor)} that transfers the query mask’s spatial information to the target view using DINOv3 $\varphi(\cdot)$’s geometry-aware feature space, for the first time enabling coordinate-based prompting across views;  
2) a \textit{Cross-View Visual Prompt Generator (V$^{2}$-Visual)} that leverages object appearance cues and refines them through a learnable mapping between views;  
3) a \textit{Multi-Expert Training} mechanism that jointly learns spatial, visual, and fused experts for complementary reasoning; and  
4) a \textit{Post-hoc Cyclic Consistency Selector (PCCS)} that adaptively selects the most reliable expert at inference based on cross-view mask consistency. Together, these components form our V$^2$-SAM, a unified segmentation framework that bridges spatial alignment and semantic association across drastically different viewpoints.

\begin{figure}[t]
    \centering
    \includegraphics[width=\linewidth]{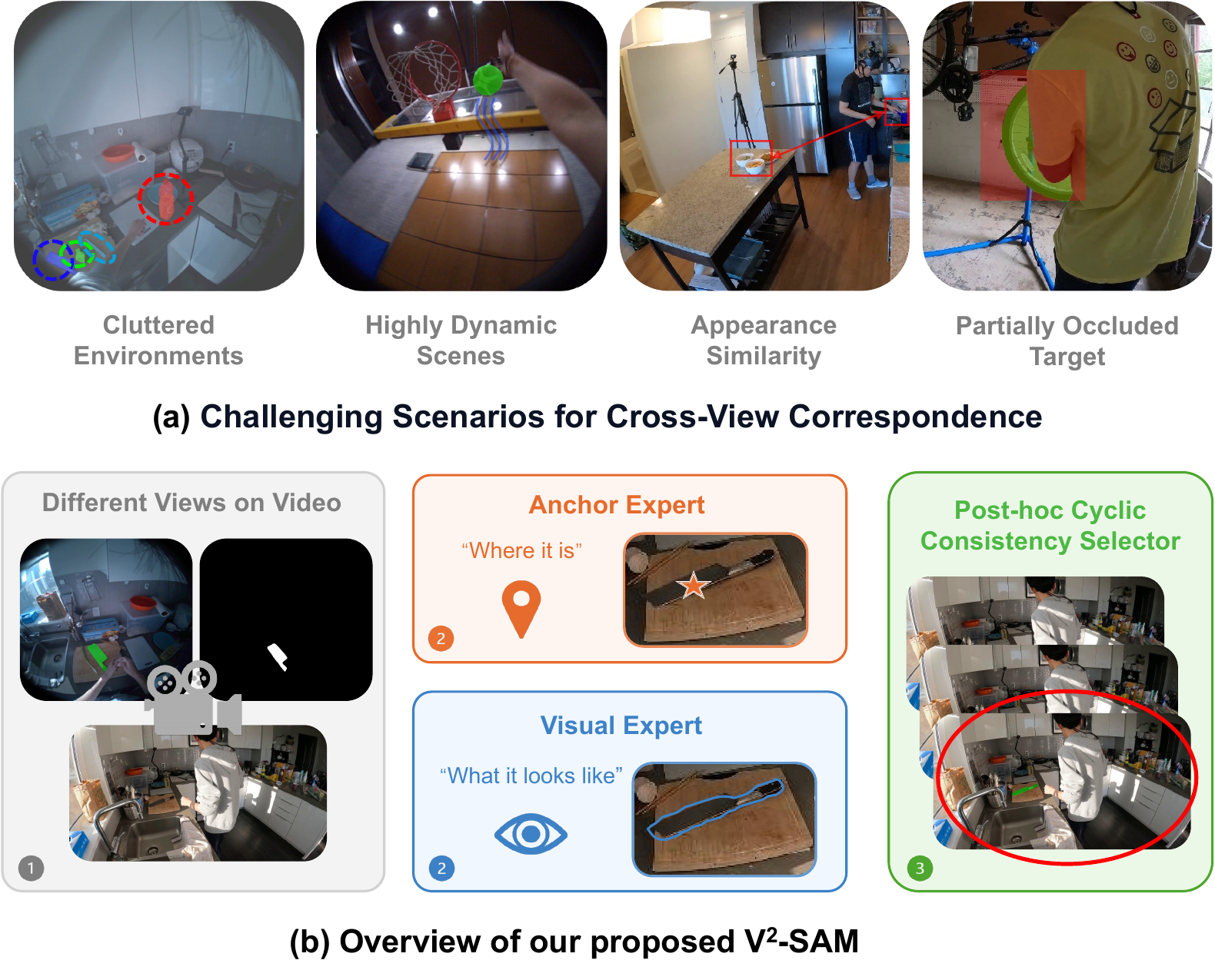}
    \caption{Cross-view correspondence challenges and our method.}
    \label{fig:supfig1}
\end{figure}

\subsection{Mixture-of-Experts in Vision}
Recent advances in the Mixture-of-Experts~\cite{he2021fastmoefastmixtureofexperttraining} (MoE) paradigm have demonstrated strong potential for scalable and adaptive visual modeling via input-dependent expert routing in computer vision~\cite{pan2025enhance, zhu2024customize}. Building on this idea, TimeExpert~\cite{Yang2025timeexpert} extends dynamic routing to spatiotemporal modeling, FlexTrack~\cite{Tan2025flextrack} introduces heterogeneous experts for adaptive computation, and SM3Det~\cite{li2024sm3det} employs sparse grid-level experts for multi-modal detection for remote sensing~\cite{pan2025locate,pan2023reducing,ma2024direction}; meanwhile, XTrack~\cite{Tan2025xtrack} and ProMoE~\cite{wei2025promoe} further improve modality fusion and routing precision. Despite these advances, existing MoE designs primarily focus on single-view or modality-level adaptation. In contrast, we propose a multi-prompt expert framework that adaptively selects spatial and visual experts to handle cross-view correspondence under drastic viewpoint and appearance changes.

\section{Challenges in Cross-View Object Correspondence}
Cross-view object correspondence in real-world environments remains highly challenging due to substantial intra-scene variations and visual ambiguity across viewpoints, as shown in Fig.~\ref{fig:supfig1}. First, \textit{cluttered scenes} with numerous overlapping objects introduce significant distractors, making it difficult to reliably localize the same instance across views. Second, \textit{highly dynamic settings}, where either the camera or objects exhibit rapid motion, lead to drastic changes in appearance, lighting, and geometry. Third, the presence of \textit{appearance-similar objects} (e.g., tools with comparable shapes and textures) often results in ambiguous matching. Finally, \textit{partial occlusions} may cause the target to be only partially visible or temporarily absent in certain views, further increasing correspondence difficulty.

To address these challenges, we propose V$^2$-SAM, a unified framework that jointly leverages (i) a \textit{Anchor Expert} to reason about object location (``where it is''), (ii) a \textit{Visual Expert} to capture fine-grained appearance cues (``what it looks like''), and (iii) a \textit{Post-hoc Cyclic Consistency Selector} to enforce cross-view agreement. Collectively, these components enable more robust and accurate cross-view correspondence in complex, real-world settings.

\begin{table}[t]
\centering
\small
\begin{tabular}{l l c c c c}
\toprule
Dataset & Subset & Split & Pairs & Masks & \# Classes \\
\midrule
\multirow{4}{*}{Ego-Exo4D} 
    & Ego2Exo & Train & 110K & 523K & $\sim$28 \\
    & Ego2Exo & Test  & 41K  & 200K & $\sim$35 \\
    & Exo2Ego & Train & 123K & 567K & $\sim$29 \\
    & Exo2Ego & Test  & 47K  & 219K & $\sim$35 \\
\midrule
\multirow{2}{*}{HANDAL-X} 
    & - & Train & 39K & 78K & 17 \\
    & - & Test  & 13K & 26K & 17 \\
\midrule
\multirow{2}{*}{DAVIS-17} 
    & - & Train & 2.8K & 12.7K & 8 \\
    & - & Test  & 1.4K & 5.3K & 5 \\
\bottomrule
\end{tabular}
\caption{Statistics of datasets used in our experiments. Ego-Exo4D is divided into directional subsets Ego2Exo and Exo2Ego.}
\label{tab:dataset_stats}
\end{table}

\section{More Implementation Details}
\subsection{Dataset Settings}

Tab.~\ref{tab:dataset_stats} provides a quantitative overview of the datasets used in our experiments. Our primary supervision comes from Ego-Exo4D, where we leverage two directional splits: \textit{Ego2Exo} and \textit{Exo2Ego}. Each direction includes both training and testing sets, totaling over 320K pairs and 1.5M masks across roughly 30 semantic categories. This large-scale paired data enables cross-view correspondence learning between egocentric and exocentric perspectives. HANDAL-X contributes an additional 52K pairs and 104K masks covering 17 categories of manipulable objects, providing object-level diversity while still maintaining a focus on real-world, robotics-relevant items. Finally, DAVIS-17 provides 4.2K pairs and 18K high-quality masks for multi-object video segmentation, with 8 training classes and 5 testing classes, serving as a benchmark for generalization in dense video segmentation.

\noindent \textbf{Ego-Exo4D.}
Ego-Exo4D contains synchronized egocentric and exocentric videos of human activities. We use two directional subsets: Ego2Exo (110K/41K train/test pairs, 523K/200K masks) and Exo2Ego (123K/47K train/test pairs, 567K/219K masks), spanning roughly 28--35 classes. These splits provide large-scale supervision for cross-view mapping.

\noindent \textbf{HANDAL-X.}
HANDAL-X contains real-world manipulable objects with 6-DoF pose labels. We use both the train (39K pairs, 78K masks) and test (13K pairs, 26K masks) splits, spanning 17 object categories. Its focus on object instances rather than semantic classes makes it well suited for object-centric manipulation tasks.

\noindent \textbf{DAVIS-2017.}
DAVIS-2017 serves as a high-quality benchmark for multi-object video segmentation. We use 2.8K training pairs (12.7K masks, 8 classes) and 1.4K testing pairs (5.3K masks, 5 classes), providing dense pixel-level masks for evaluating generalization.

\begin{table}[t]
\centering
\small
\begin{tabular}{l l}
\toprule
\multicolumn{1}{l}{Hyperparameters} & Value\\
\midrule
Batch size (per device) & 16 \\
Gradient accumulation steps & 4 \\
Effective batch size & 64 \\
Training epochs & 24 \\
Validation frequency & every 2 epochs \\
\midrule
Optimizer & AdamW \\
Learning rate & $4\times10^{-5}$ \\
Adam betas & $(0.9,\, 0.999)$ \\
Weight decay & 0.05 \\
Gradient clipping norm & 1.0 \\
Precision & bfloat16 (mixed precision) \\
\midrule
Warm-up strategy & linear warm-up \\
Warm-up ratio & 0.05 of total epochs \\
Learning rate schedule & cosine annealing \\
\bottomrule
\end{tabular}
\caption{Training hyperparameters used in our experiments.}
\label{tab:hyperparams}
\end{table}

\subsection{Training Hyperparameters} 
A summary of our training hyperparameters is provided in Tab.~\ref{tab:hyperparams}.  We train our model for 24 epochs with an effective batch size of 64, obtained by using a per-device batch size of 16 and four gradient accumulation steps. The optimization is performed using the AdamW optimizer with a learning rate of $4\times10^{-5}$, $\beta_1=0.9$, $\beta_2=0.999$, and a weight decay of 0.05. We employ mixed-precision training with \texttt{bfloat16} and dynamic loss scaling, and apply gradient clipping with a maximum norm of 1.0 to stabilize training. The learning rate follows a linear warm-up schedule for the first 5\% of epochs, followed by a cosine annealing policy with a minimum learning rate of zero. Validation is conducted every two epochs.

\subsection{Model settings}
The model configuration is summarized in Tab.~\ref{tab:modelsettings}. Our model is built upon the V$^{2}$SAM framework with a fully trainable SAM2 decoder and a grounding encoder. We initialize the model with pretrained weights and resize all input images to 1024 pixels using a direct resizing operator. The training objective combines a sigmoid-activated binary cross-entropy loss and a naive Dice loss, weighted by 2.0 and 0.5 respectively. We additionally employ point-sampled supervision to enhance mask quality. The entire system is trained using a length-grouped sampler and a video-aware collation strategy to accommodate variable-length multimodal data.

\section{More Experiments}

\subsection{Ablation on Submodule}

Tab.~\ref{tab:ablation} presents the ablation results of the proposed components, including the two submodules of V$^{2}$-Visual (Semantic Mapping and Spatial Mapping), the associated losses $\mathcal{L}_{v}$ and $\mathcal{L}_{s}$, and the V$^{2}$-Anchor. Each component contributes positively to overall performance, while V$^{2}$-Anchor yields the greatest improvement.

Comparing the first two rows, the configuration with Semantic Mapping and $\mathcal{L}_{v}$ slightly outperforms the one using Spatial Mapping and $\mathcal{L}_{s}$, suggesting that enforcing semantic consistency plays a more crucial role than spatial alignment when transferring visual cues across views. When both mapping modules are jointly enabled with their respective losses (third row), we observe further gains in overall IoU, demonstrating that semantic and spatial constraints are complementary and collaboratively enhance cross-view coherence. The most substantial improvement is observed after introducing V$^{2}$-Anchor (fourth row). Acting as a stable cross-view reference, V$^{2}$-Anchor significantly strengthens ego–exo alignment, leading to relative improvements of 30.2\% (Ego2Exo), 4.7\% (Exo2Ego), and 15.7\% (Total). These results indicate that anchoring the visual prompts effectively reduces cross-view ambiguity and stabilizes the mapping process.

Overall, the ablation study highlights the synergy among the proposed components: the V$^{2}$-Visual module establishes the foundation for consistent cross-view representation, while V$^{2}$-Anchor further amplifies this effect, enabling the full framework to achieve the best performance.

\begin{table}[t]
\centering
\small
\begin{tabular}{l l}
\toprule
\multicolumn{1}{l}{Model Settings} & Value\\
\midrule
Backbone model &  SAM2 \\
Decoder & SAM2 decoder (trainable) \\
Encoder 1 & SAM2 encoder \\
Encoder 2 & DINOv3 encoder \\
Pretrained checkpoint & pretrained SAM2 \\
Image pre-processing & direct resizing to 1024 px \\
Use point-sampled supervision & Yes \\
\midrule
Binary Cross-Entropy loss weight & 2.0 \\
Dice loss weight & 0.5 \\
Dice activation & sigmoid + activation \\
Dice formulation & naive Dice variant \\
\bottomrule
\end{tabular}
\caption{Model configuration and loss functions.}
\label{tab:modelsettings}
\end{table}

\subsection{Ablation on V$^2$-Anchor}
% 1 5 10 30 all (point for anchor expert)
% w/o mask vs. with mask
Tab.~\ref{tab:anchor_point_ablation} reports the impact of varying the number of sparse anchor points in the V$^{2}$-Anchor Expert. In the Ego→Exo setting, performance monotonically degrades as anchors become denser, dropping from 38.7 IoU with a single anchor to 32.2, 28.7, and 21.8 with 5, 10, and 30 anchors. A similar trend is observed in the Exo→Ego direction, where the 1-anchor configuration achieves the best result (41.6 IoU), while denser anchors significantly reduce accuracy (e.g., 18.3 and 18.5 IoU with 10 and 30 anchors).

\begin{table}[h]
\centering
\setlength{\tabcolsep}{6pt}
\renewcommand{\arraystretch}{1.15}
\resizebox{0.78\columnwidth}{!}{
\begin{tabular}{lcccc}
\toprule
Setting & 1 pt & 5 pts & 10 pts & 30 pts \\
\midrule
Ego$\rightarrow$Exo & 38.7 & 32.2 & 28.7 & 21.8 \\
Exo$\rightarrow$Ego & 41.6 & 26.1 & 18.3 & 18.5 \\
\bottomrule
\end{tabular}}
\caption{Ablation of sparse anchor point count in the V$^{2}$-Anchor Expert. We report IoU under Ego$\rightarrow$Exo and Exo$\rightarrow$Ego settings using 1, 5, 10, and 30 correspondence points.}
\label{tab:anchor_point_ablation}
\end{table}

These results suggest that a minimal set of high-confidence cross-view correspondences provides stronger supervisory constraints, whereas denser anchor distributions may introduce view-specific noise or limit generalization. Overall, sparse anchoring proves more effective for cross-view alignment across both transfer directions.

% \begin{table}[t]
% \centering
% \setlength{\tabcolsep}{6pt}
% \renewcommand{\arraystretch}{1.15}
% \resizebox{0.9\columnwidth}{!}{
% \begin{tabular}{lccccc}
% \toprule
% Setting & 1 pt & 5 pts & 10 pts & 30 pts & All pts \\
% \midrule
% Ego$\rightarrow$Exo & 38.7 & 32.2 & 28.7 & 21.8 & 32.2 \\
% Exo$\rightarrow$Ego & 41.6 & 26.1 & 18.3 & 18.5 & 32.6 \\
% \bottomrule
% \end{tabular}}
% \caption{Ablation of sparse anchor point count in the V$^{2}$-Anchor Expert. We report mIoU under Ego$\rightarrow$Exo and Exo$\rightarrow$Ego settings using 1, 5, 10, 30, and all correspondence points.}
% \label{tab:anchor_point_ablation}
% \end{table}

% =====Ablation on the proposed modules in VP Expert.========
\begin{table*}[t]
  \centering
  \setlength{\tabcolsep}{6pt}
  \renewcommand{\arraystretch}{1.15}
  \resizebox{0.85\linewidth}{!}{
  \begin{tabular}{l l c c c c}
    \toprule
    Decoders & Selector & Ego2Exo IoU$\uparrow$ & Exo2Ego IoU$\uparrow$ & Runtime (ms/sample)$\downarrow$ & FLOPs (G/sample)$\downarrow$ \\
    \midrule
    \multirow{2}{*}{A{+}B}
      & Cycle-Mask (Prior)        &  {42.60} &  {46.73} & {620} &  {2188.58 GFLOPs}\\
      & \textbf{Cycle-Points (Ours)} &  {\textbf{42.71}} &  {\textbf{48.17}} &  {\textbf{510}} &  {\textbf{2173.58 GFLOPs}} \\
    \midrule
    \multirow{2}{*}{A{+}B{+}C}
      & Cycle-Mask (Prior)        &  {46.27} &  {49.43} &  {820} &  {2207.13 GFLOPs} \\
      & \textbf{Cycle-Points (Ours)} &  {\textbf{46.31}} &  {\textbf{49.61}} &  {\textbf{760}} &  {\textbf{2184.63 GFLOPs}} \\
    \bottomrule
  \end{tabular}}
\caption{Ablation on the Post-hoc Cyclic Consistency Selector in \textbf{Ego$\leftrightarrow$Exo} correspondence. 
We compare \textbf{Cycle-Points (Ours)} with the mask-based \textbf{Cycle-Mask (Prior)} 
on two decoder combinations: A{+}B (Anchor+Visual) and A{+}B{+}C (Anchor+Visual+Fusion). 
}
  \label{tab:cycle_selector_ablation}
\end{table*}

\begin{table*}[t]
    \centering
    \resizebox{0.85\linewidth}{!}{
    \begin{tabular}{c cccc ccc}
        \toprule
        \makecell{V$^{2}$-Visual\\Semantic Mapping} & \makecell{V$^{2}$-Visual\\Spatial Mapping} & $\mathcal{L}_{v}$ & $\mathcal{L}_{s}$ & V$^{2}$-Anchor &
        \makecell{Ego2Exo\\IoU$\uparrow$} &
        \makecell{Exo2Ego\\IoU$\uparrow$} &
        \makecell{Total\\IoU$\uparrow$} \\
        \midrule
        \checkmark & \checkmark & - & \checkmark & - & {34.19} & {45.16} & {39.68} \\
        \checkmark & - & \checkmark & - & - & {34.98} & {46.46} & {40.72} \\
        \checkmark & \checkmark & \checkmark & \checkmark & - & {36.17} & {46.63} & {41.40} \\
        \checkmark & \checkmark & \checkmark & \checkmark & \checkmark & {44.51} & {47.29} & {45.90} \\
        \multicolumn{5}{c}{Relative Gain $\%$ of x with respect to y $\frac{(x-y)}{y}$}& 
        \textcolor{ForestGreen}{+30.2$\%$} & 
        \textcolor{ForestGreen}{+4.7$\%$} & 
        \textcolor{ForestGreen}{+15.7$\%$} \\
        \bottomrule
    \end{tabular}
    }
    \caption{{Ablation study of the proposed modules V$^{2}$-Visual, V$^{2}$-Anchor, $\mathcal{L}_{v}$ and $\mathcal{L}_{s}$ on the test set. The Cross-View Visual Prompt Generator (V$^{2}$-Visual) consists of two submodules, Semantic Mapping and Spatial Mapping, which correspond to Semantic Constraint and Spatial Constraint, respectively.}}
    \label{tab:ablation}
\end{table*}

% =====Ablation on the Post-hoc Cyclic Consistency Selector.========
\subsection{Ablation on the PCCS}
Tab. \ref{tab:cycle_selector_ablation} compares our PCCS with the prior Cycle-Mask Selector, which reconstructs query-view masks for cyclic validation.
Instead of predicting masks, our method directly measures geometric consistency between predicted query points and sampled reference points from the raw prompt mask, selecting the expert with the smallest cyclic distance.
Despite its simpler design, Cycle-Points achieves comparable or even higher accuracy while notably reducing computation.
For the two-expert setup (A+B), it improves Exo→Ego IoU by +1.4 and shortens runtime by 110 ms per sample.
With three experts (A+B+C), it maintains accuracy while reducing both latency and FLOPs.

These results demonstrate that expert selection can be effectively driven by point-level cyclic consistency alone, providing a lightweight yet equally reliable alternative to mask-based cyclic reasoning.

\subsection{More Visual Analytics.}

% Fig1: ref sam vs our v2-sam
%%%%%%%%%%%%%%%%%%%%%%%%%%%%%% 2组 exo2ego ego2exo
% 3-4组 exo2ego ego2exo

% Fig2: visual expert (3个专家的对比+PCCS，找更多的例子）
%% 1组 exo2ego ego2exo

% Fig3: PCCS是如何work的
%% 3组 exo2ego ego2exo （可视化跟label的mask的中心距离-点和距离）
%% 在overlay的图像（query和target）上进行3个专家点的可视化DINOv3后循环一致性的点的位置，需要两个视角

% Fig4: Anchor是如何work的
% 1组例子ego2exo exo2ego

% Fig5: before VPMatcher vs after VPMatcher
% 1组例子ego2exo exo2ego

% Fig6: handal and Davis
% 最后做
%%%%%%%%%%% TODO：Handal Fintune and 更多的Ref-SAM 的方法
%%%%%%%%%%%% TODO：val的(更优先，每个场景用最好的专家推）

\begin{figure*}[t]
    \centering
    \begin{subfigure}[t]{0.49\linewidth}
        \centering
        \includegraphics[width=\linewidth]{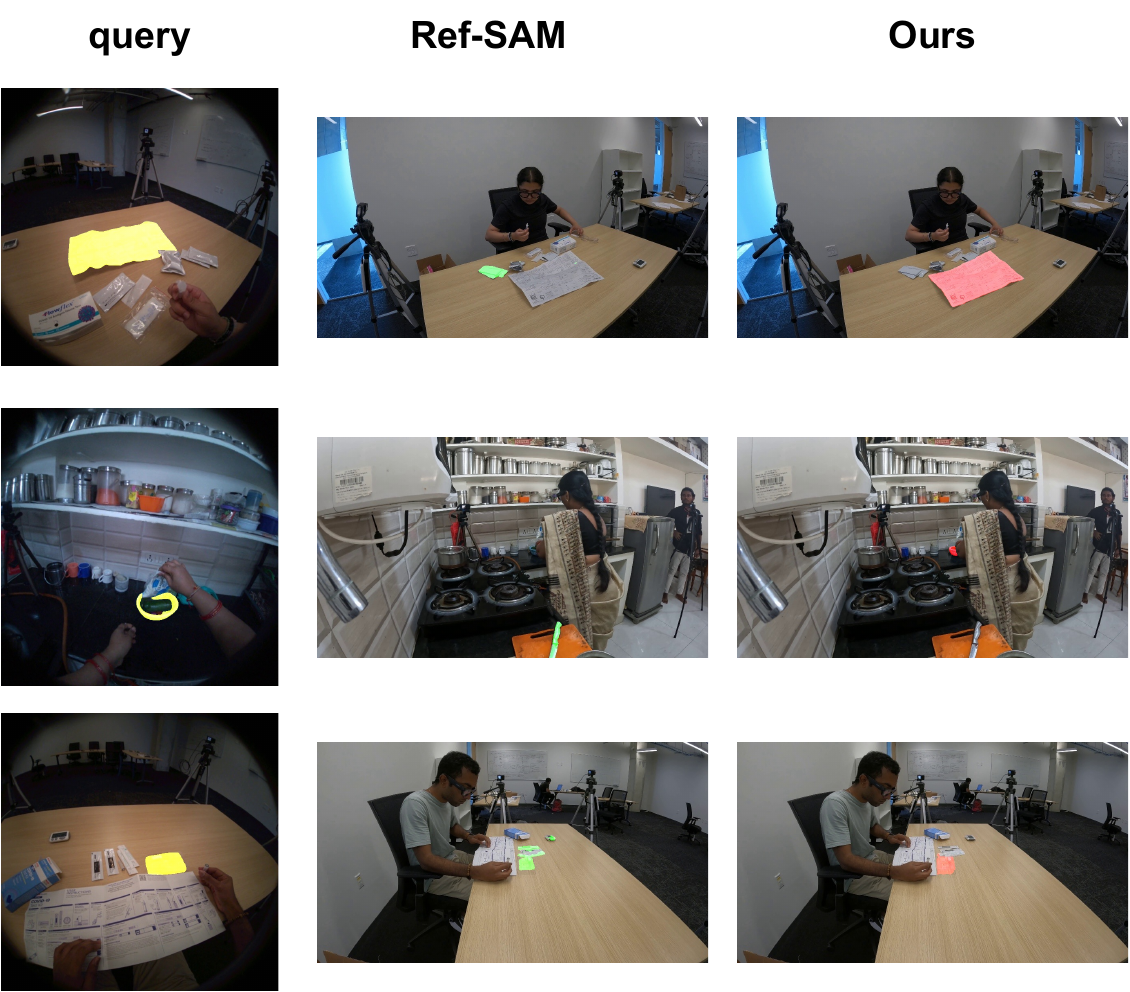}
        \caption{Ego2Exo.}
        \label{fig:eg2ex}
    \end{subfigure}
    \hfill
    \begin{subfigure}[t]{0.44\linewidth}
        \centering
        \includegraphics[width=\linewidth]{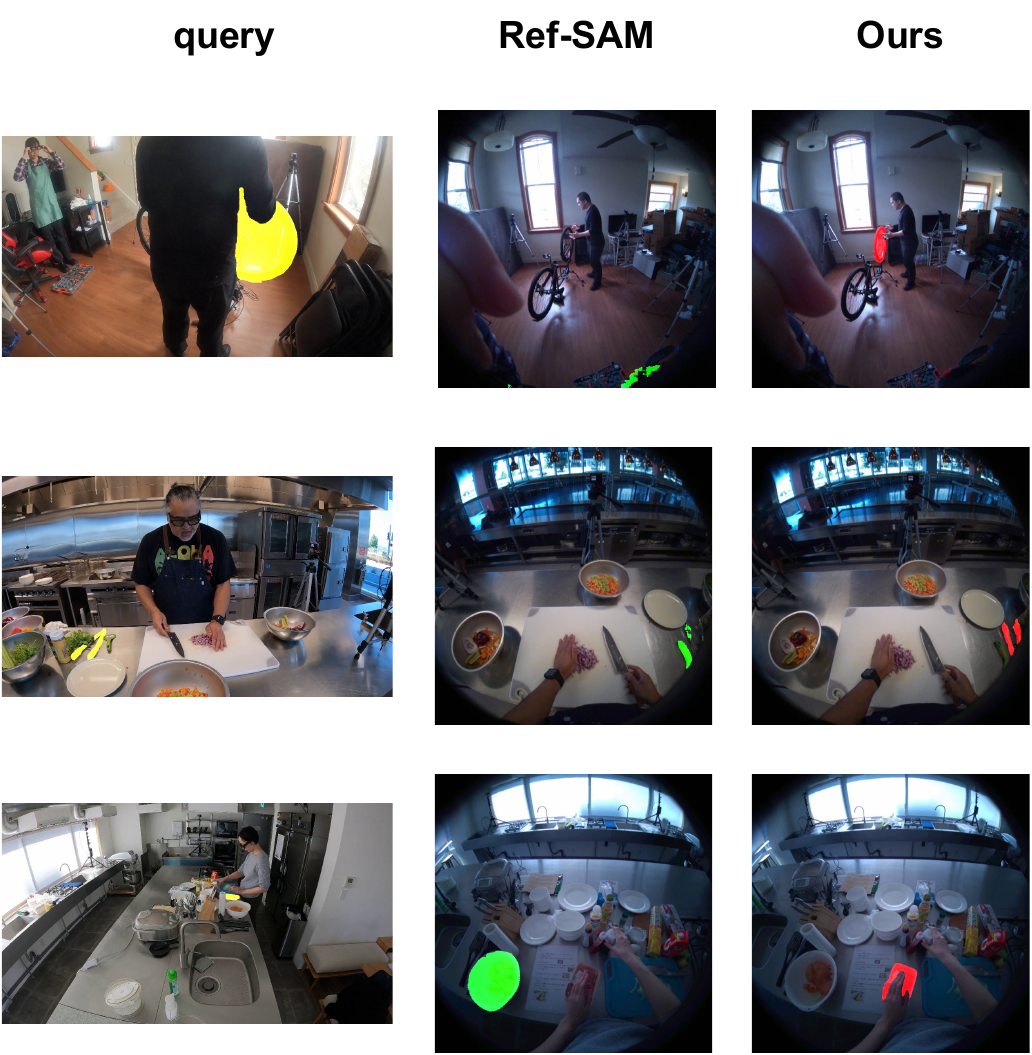}
        \caption{Exo2Ego.}
        \label{fig:ex2eg}
    \end{subfigure}
    \caption{Qualitative comparison with Ref-SAM on the Ego-Exo4D dataset under two cross-view settings. The left column in each subfigure shows the query image, followed by predictions from Ref-SAM and our method. Our approach produces more accurate and consistent cross-view localization across both Ego2Exo and Exo2Ego scenarios.}
    \label{fig:crossview}
\end{figure*}

\noindent \textbf{Ref-SAM VS. V$^2$-SAM.}
Fig.~\ref{fig:eg2ex} and Fig.~\ref{fig:ex2eg} show the visual quality comparison of Ref-SAM and our proposed V$^2$-SAM. Ref-SAM often fails to accurately localize target objects in cluttered scenes or when multiple similar objects are present. In contrast, our method generates precise and consistent predictions across both directions, illustrating superior robustness and cross-view generalization. These results further confirm that our method effectively bridges the viewpoint gap between egocentric and exocentric observations.

\begin{figure*}[t]
    \centering
    \includegraphics[width=0.98\linewidth]{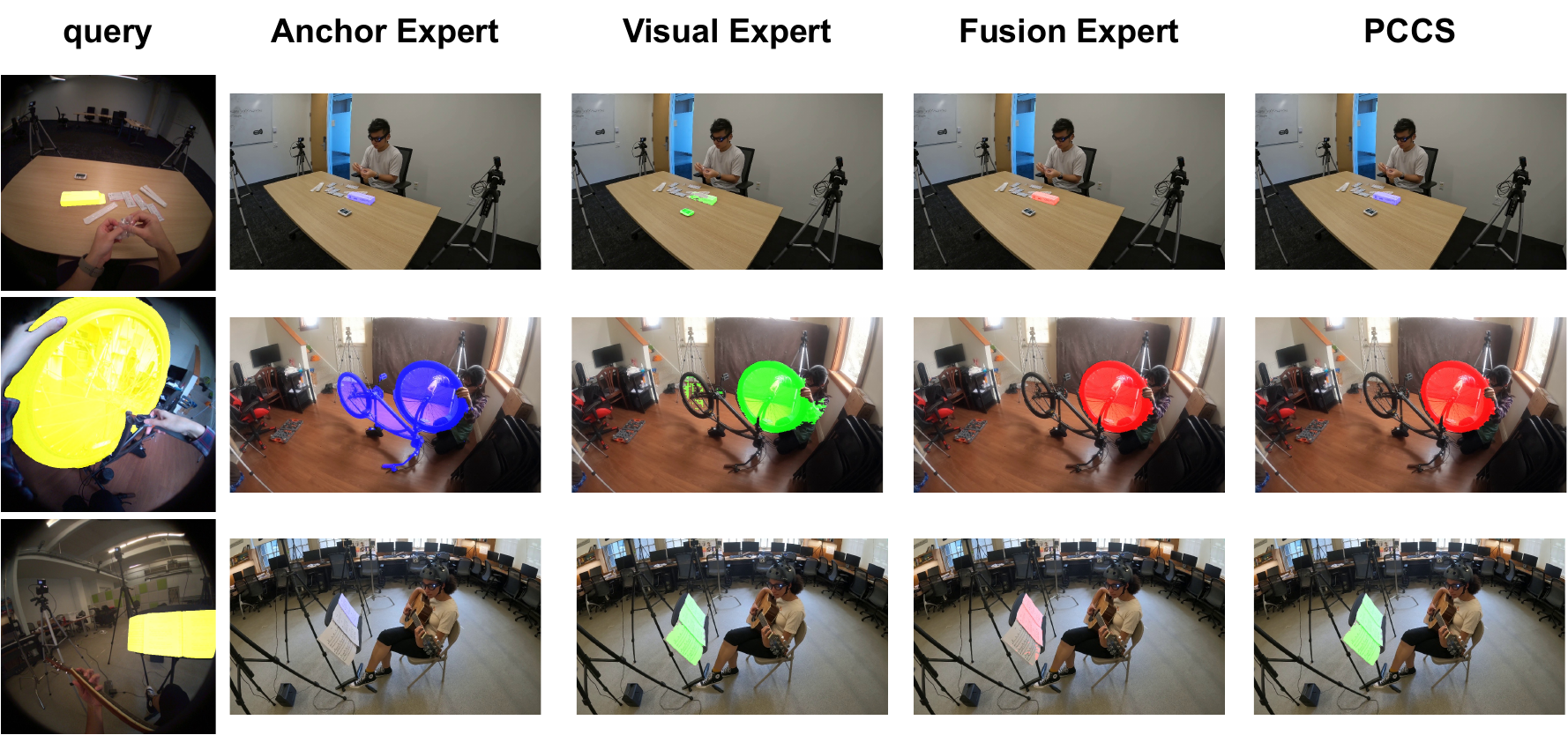}
    \caption{Comparison of selection results among individual experts and the PCCS on the Ego2Exo task. Each expert demonstrates varying strengths in interpreting specific first-person perspectives, while the PCCS leverages consensus across experts to improve overall selection consistency.}
    \label{fig:supfig3}
\end{figure*}

\begin{figure*}[t]
    \centering
    \includegraphics[width=0.98\linewidth]{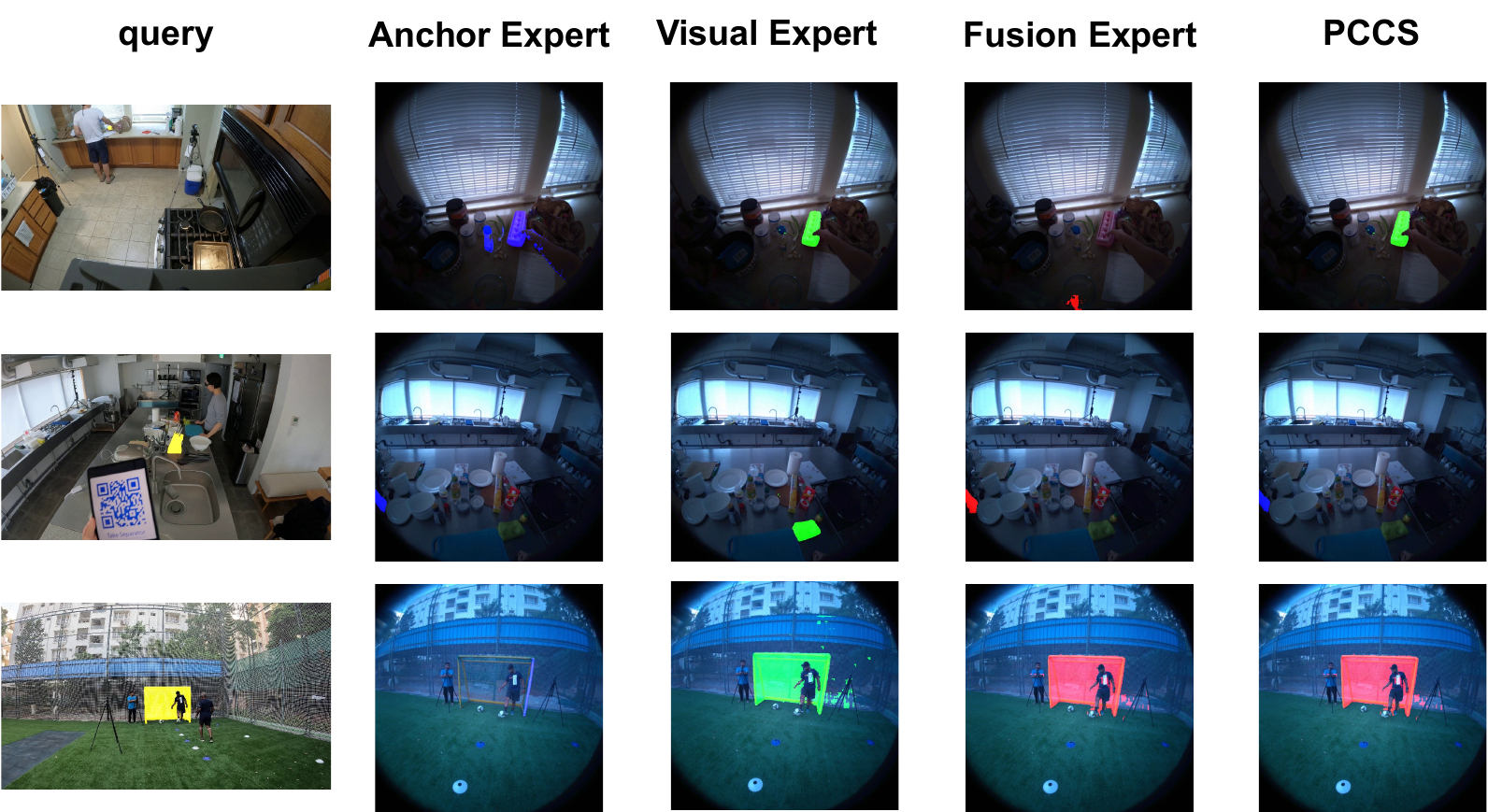}
    \caption{Comparison of selection results among individual experts and the PCCS on the Exo2Ego task. Experts show diverse interpretative preferences for third-person viewpoints, and the PCCS consolidates these judgments to yield more robust and balanced decisions.}
    \label{fig:supfig4}
\end{figure*}

\noindent \textbf{Different Experts.}
Fig.~\ref{fig:supfig3} and Fig.~\ref{fig:supfig4} present the comparative results obtained from multiple experts and the PCCS for the Ego2Exo and Exo2Ego tasks, respectively. The results demonstrate that combining expert assessments with the PCCS has a substantial influence on the final selection process. Moreover, different experts exhibit complementary strengths in understanding distinct scenario types, underscoring the benefit of aggregating diverse human expertise with system-level decision-making.

\begin{figure*}[t]
    \centering
    \includegraphics[width=0.9\linewidth]{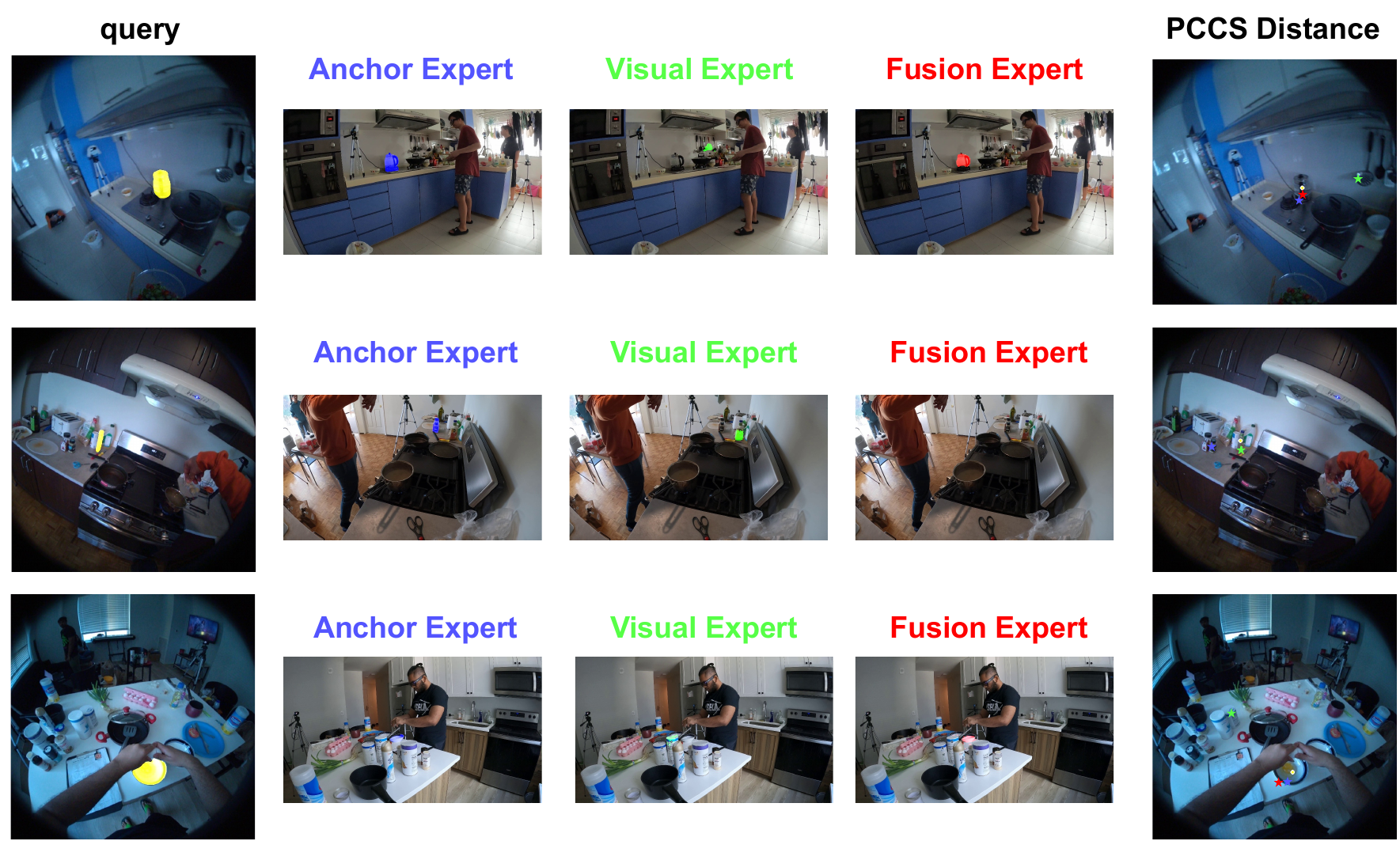}
    \caption{Ego2Exo Analysis. We quantify alignment by measuring the distance between the predicted locations of the \textcolor{blue}{Anchor Expert}, \textcolor{green}{Visual Expert}, and \textcolor{red}{Fusion Expert} (colored accordingly) and the ground-truth query-mask annotations. Benefiting from integrating heterogeneous expert preferences, PCCS selects the expert whose prediction is closest to the annotation centroid.}
    \label{fig:supfig5}
\end{figure*}

\begin{figure*}[t]
    \centering
    \includegraphics[width=0.9\linewidth]{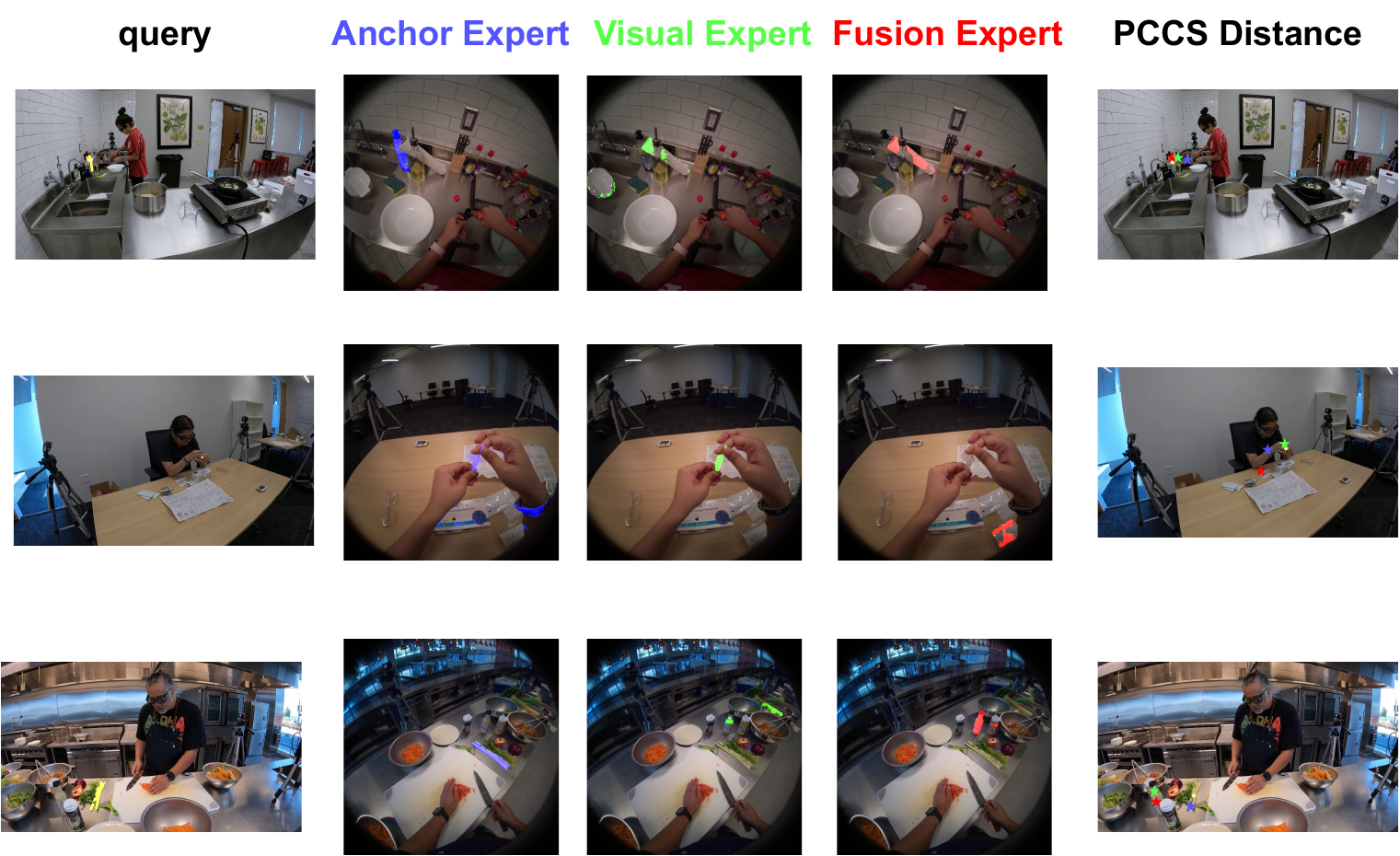}
    \caption{Exo2Ego Analysis. We quantify alignment by measuring the distance between the predicted locations of the \textcolor{blue}{Anchor Expert}, \textcolor{green}{Visual Expert}, and \textcolor{red}{Fusion Expert} (colored accordingly) and the ground-truth query-mask annotations. Benefiting from integrating heterogeneous expert preferences, PCCS selects the expert whose prediction is closest to the annotation centroid.}
    \label{fig:supfig6}
\end{figure*}

\noindent \textbf{PCCS Consensus Analysis.}
Fig.~\ref{fig:supfig5} and Fig.~\ref{fig:supfig6} analyze the mechanism of PCCS under the Ego2Exo and Exo2Ego settings. Rather than evaluating task performance, we aim to understand how PCCS forms consensus. We measure the L2 distance of different experts' prediction results and the labeled location of query masks, providing a quantitative assessment of alignment. Results indicate that individual experts exhibit distinct biases across scenarios, whereas PCCS aggregates these cues and consistently converges toward selections closer to the annotation space.

% \noindent \textbf{VPMatcher Consensus Analysis.}

\begin{figure*}[t]
    \centering
    \includegraphics[width=0.9\linewidth]{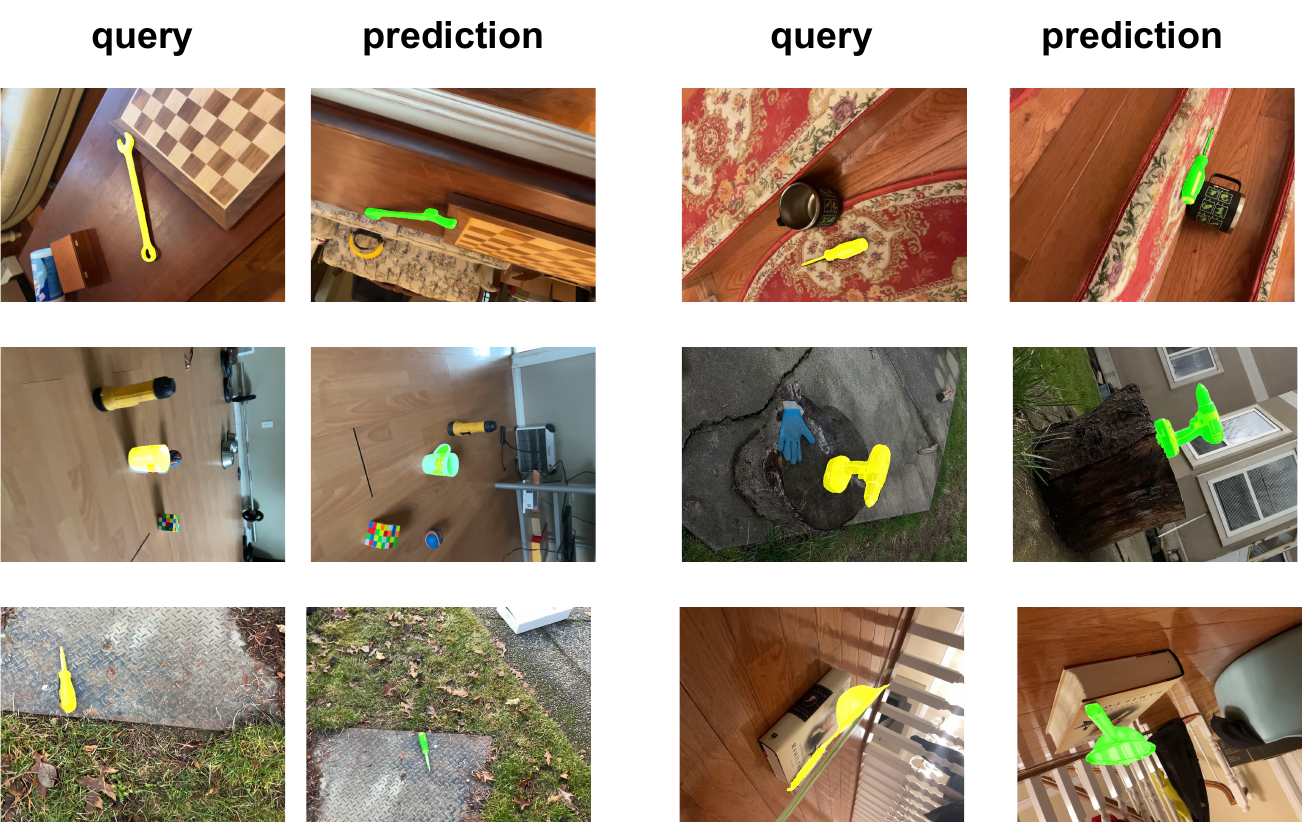}
    \caption{Visualization of the prediction results of our V$^{2}$-SAM method on the HANDAL-X dataset.}
    \label{fig:supfig7}
\end{figure*}

\begin{figure*}[t]
    \centering
    \includegraphics[width=0.9\linewidth]{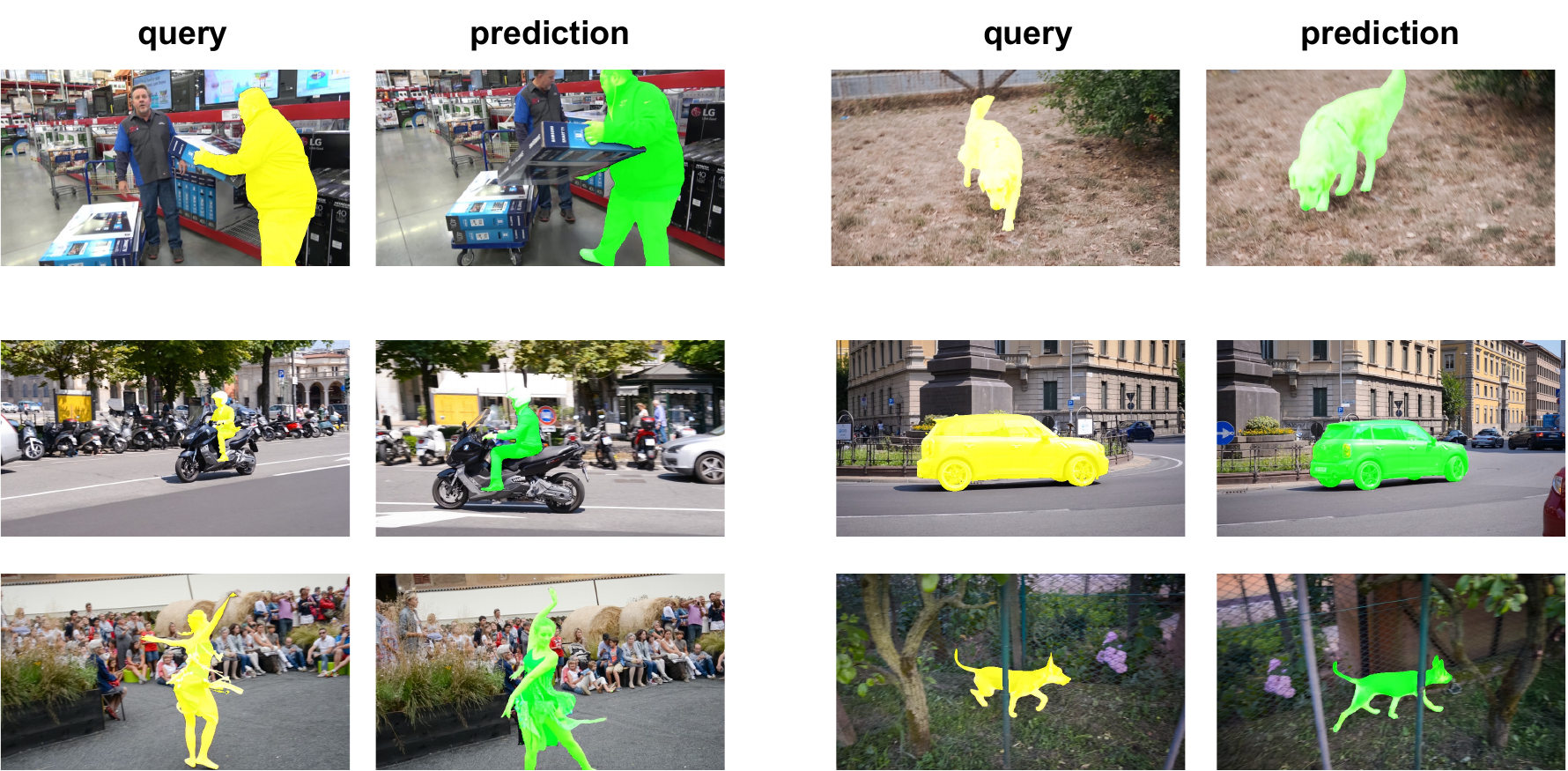}
    \caption{Visualization of the prediction results of our V$^{2}$-SAM method on the DAVIS-17 dataset.}
    \label{fig:supfig8}
\end{figure*}

\noindent \textbf{Visualization of HANDAL-X and DAVIS-17.}
Fig.~\ref{fig:supfig7} and Fig.~\ref{fig:supfig8} present qualitative results of our V$^{2}$-SAM method on the HANDAL-X and DAVIS-17 datasets, respectively. Across both indoor and outdoor scenes, V$^{2}$-SAM accurately localizes objects given only a single query frame, demonstrating strong consistency between query masks and predicted masks across subsequent viewpoints. On HANDAL-X, the model robustly segments small, elongated tools under large appearance variations and background clutter. On DAVIS-17, V$^{2}$-SAM generalizes to complex scenes involving articulated motion, occlusion, and diverse object categories, while maintaining precise object boundaries and temporal coherence. These visualizations validate that V$^{2}$-SAM effectively transfers query-driven segmentation cues across scenes and object instances.

% \begin{figure}[t]
%     \centering
%     \includegraphics[width=0.98\linewidth]{IMAGES/SupFig9.pdf}
%     \caption{Qualitative comparison on the Ego-Exo4D dataset under the Ego2Exo setting. The left column shows the query image, followed by predictions from Ref-SAM and our method.}
%     \label{fig:supfig9}
% \end{figure}

% \begin{figure}[t]
%     \centering
%     \includegraphics[width=0.98\linewidth]{IMAGES/SupFig10.pdf}
%     \caption{Qualitative comparison on the Ego-Exo4D dataset under the Exo2Ego setting. The left column shows the query image, followed by predictions from Ref-SAM and our method.}
%     \label{fig:supfig10}
% \end{figure}

% \begin{figure}
%     \centering
%     \includegraphics[width=0.99\linewidth]{IMAGES/Anchor.pdf}
%     \caption{Comparative Performance of Anchor, Visual, and Fusion Experts Across Scenes.}
%     \label{tab:expert-performance}
% \end{figure}